\documentclass[final,12pt]{clear2023} 

\title{Causal Discovery with Score Matching on Additive Models with Arbitrary Noise}

\usepackage{mathtools}      
\usepackage{times}
\usepackage{graphicx} 
\usepackage{centernot}
\usepackage{tikz}
\usepackage[font=small]{caption}
\usepackage{subcaption}
\usepackage{booktabs}
\usepackage{multirow}
\usepackage{setspace} 
\usepackage{mathtools}
\usepackage{enumitem} 

\newcommand{\Xminus}[1]{\mathbf{X}_{\mathsmaller{\setminus} \{#1\}}} 
\newcommand{\R}{\mathbb{R}} 
\newcommand{\Var}{\operatorname{Var}} 
\newcommand{\Mu}{\operatorname{\mathbf{E}}} 
\newcommand{\Parents}{\operatorname{PA}} 
\newcommand{\Child}{\operatorname{CH}} 
\newcommand{\parents}{\operatorname{pa}} 
\newcommand{\child}{\operatorname{ch}} 
\newcommand{\Desc}{\operatorname{DE}} 
\newcommand{\Nondesc}{\operatorname{ND}} 
\newcommand{\Rsub}{\mathsmaller{R_i}} 

\DeclarePairedDelimiterX{\expectarg}[1]{[}{]}{%
  \ifnum\currentgrouptype=16 \else\begingroup\fi
  \activatebar#1
  \ifnum\currentgrouptype=16 \else\endgroup\fi
}
\newcommand{\innermid}{\nonscript\;\delimsize\vert\nonscript\;}
\newcommand{\activatebar}{%
  \begingroup\lccode`\~=`\|
  \lowercase{\endgroup\let~}\innermid 
  \mathcode`|=\string"8000
}

\clearauthor{%
 \Name{Francesco Montagna}\thanks{Work has been partially carried out during an internship at Amazon Web Services, Tubingen} \Email{francesco.montagna@edu.unige.it}\\
    \addr MaLGa-DIBRIS, Università di Genova
 \AND
  \Name{Nicoletta Noceti} \Email{nicoletta.noceti@unige.it}\\
  \addr MaLGa-DIBRIS, Università di Genova
 \AND
  \Name{Lorenzo Rosasco} \Email{lrosasco@mit.edu}\\
  \addr MaLGa-DIBRIS, Università di Genova\\
  MIT, CBMM\\
  Istituto Italiano di Tecnologia
 \AND
  \Name{Kun Zhang} \Email{kunz1@cmu.edu}\\
  \addr Carnegie Mellon University\\
  MBZUAI
 \AND
 \Name{Francesco Locatello} \Email{locatelf@amazon.com}\\
 \addr AWS
}

\begin{document}
\maketitle

\begin{abstract}
\looseness-1Causal discovery methods are intrinsically constrained by the set of assumptions needed to ensure structure identifiability. 
Moreover additional restrictions are often imposed in order to simplify the inference task: this is the case for the Gaussian noise assumption on additive nonlinear models, which is common to many causal discovery approaches. In this paper we show the shortcomings of inference under this hypothesis, analyzing the risk of edge inversion under violation of Gaussianity of the noise terms. Then, we propose a novel method for inferring the topological ordering of the variables in the causal graph, from data generated according to an additive nonlinear model with a generic noise distribution. This leads to NoGAM (Not only Gaussian Additive noise Models), a causal discovery algorithm with a minimal set of assumptions and state of the art performance, experimentally benchmarked on synthetic data.
\end{abstract}


\section{Introduction}
\looseness-1Inferring cause-effect relationships from observational data is a central goal of causality research,
as it enables formal reasoning about interventions on a system (\cite{peters_2017}, \cite{Pearl09}) when these are expensive, unethical, or even impossible to perform.
Structure identifiability results posit limits to what part of the causal graph can be inferred from pure observations from the joint distribution, and provides formal guidelines on which assumptions are needed to fully identify the causal graph underlying the data. Traditional causal discovery methods usually are limited to identify Markov equivalence classes (\cite{cd_review_glymour}), which is the case for PC, FCI (\cite{Spirtes2000}) and GES (\cite{chickering_2003}). More recently, methods based on properly defined Structural Causal Models (SCMs) have been proposed to distinguish the correct graph underlying the observed data, by mean of additional assumptions on the functional class of the SCM: \cite{hoyer2009ANM} and \cite{zhang2009PNL} show that nonlinear additive noise models typically yield an identifiable setting. This is the case for SCORE (\cite{rolland_2022}) and CAM (\cite{B_hlmann_2014}) that, under the assumption of Gaussian disturbances, output a unique and asymptotically consistent graph as result of the inference process. Under the condition of identifiable nonlinear additive models, \cite{peters_2014_identifiability} and \cite{mooij2009} show how to exploit independence of the estimated residuals to infer causal effects without restrictions on the noise distributions. Their methods are limited by the use of conditional independence testing, which is hard to perform (\cite{shah2018_ci_test}). Closer to our work, \cite{bloebaum18a} compare regression errors to distinguish cause and effect, but in the restricted setting of bivariate models.
Other methods such as \cite{grandag_19} and \cite{NEURIPS2018_notears} formulate a continuous optimization problem which results in a unique directed acyclic graph (DAG).

\vspace{.7em}
\looseness-1In general, identifiability results require assumptions in  order to infer the causal structure from observational data with theoretical guarantees. The shortcoming of this approach is that constraints in the form of assumptions reduce the scope of applicability of an algorithm. Instead it would be desirable to have methods working on a broad range of problems under different conditions, ideally showing a certain degree of robustness regarding violations of the model hypothesis. The strength of this viewpoint is manifest in deep learning practice, where the dominant approach is to apply algorithms that work on the task of interest, independently of the violation of the underlying assumptions. The motivation behind this paper is to provide a causal discovery tool in between these philosophies, by removing  (from the identifiability perspective) unnecessary assumptions frequently made by some of the most prominent computational methods available. With this goal in mind, we design an algorithm for the inference of the causal graph underlying an additive nonlinear model with generic noise terms, removing the common hypothesis of Gaussian distributions. This constraint removal broadens the scope of applicability of principled causal discovery, providing a state of the art method to practitioners interested in theoretical guarantees and operating in critical settings where the validity of the Gaussian noise assumption is hard to verify.

\vspace{.7em}
\looseness-1The rest of the paper is organized as follow: Section \ref{sec:background} provides an overview of the model under study, and a definition of the problem at hand; Section \ref{sec:gauss_hp} analyzes the risk of inferring inversed edges under violation of the Gaussian noise assumption; Section \ref{sec:method} introduces a theoretically principled method to find the topological ordering of a causal graph by iteratively identifying its leaf nodes: in particular we prove an important relation between the score function (i.e. the gradient of the log-likelihood) and the residuals' estimators; Section \ref{sec:algorithm} defines NoGAM\footnote{The code for NoGAM is available as part of the DoDiscover library \url{https://www.pywhy.org/dodiscover/dev/index.html}}, an algorithm for inference of the causal graph from the data; Section \ref{sec:experiments} is an overview of the experimental performance of such method with respect to classical and state of the art benchmarks.

\section{Background knowledge}\label{sec:background}
\paragraph{Model definition} Let $\mathbf{V} = \{1, \ldots, d\}$ be the vertices of a directed acyclic graph $\mathcal{G}$, and $\mathbf{X} \in \mathbb{R}^d$ be a set of random variables generated according to the Structural Causal Model (SCM)
\begin{equation}
        X_i \coloneqq f_i(\Parents_i(\mathbf{X})) + N_i, \;\; \forall \: i \in \mathbf{V} \,,
\label{eq:scm}
\end{equation}
where $\Parents_i(\mathbf{X})$ is the vector of parents of node $i$ in the graph $\mathcal{G}$. Model \eqref{eq:scm} is known as the nonlinear Additive Noise Model (ANM, \cite{hoyer2009ANM}). We assume causal mechanisms $f_i$ to be nonlinear functions continuously differentiable, and causal minimality to be satisfied. Noise terms $N_i \in \R$ are continuous random variables with density $p_i(N_i)$, mean $\mu_i = 0$ and variance $\sigma_i^2 > 0$. We assume them to be independent such that their joint distribution is $p_\mathbf{N}(\mathbf{N}) = \prod_i p_i(N_i)$. Under these assumptions the model \eqref{eq:scm} induces a joint distribution $p_{\mathbf{X}}(\mathbf{X})$ which is Markov with respect to $\mathcal{G}$, such that it admits the following factorization:
\begin{equation}
    p_{\mathbf{X}}(\mathbf{X}) = \prod_i^d p_i(X_i \mid \Parents_i(\mathbf{X})) \:,
\label{eq:factorization}
\end{equation}
where with an abuse of notation we distinguish the marginal $p_i(N_i)$ from $p_i(X_i \mid \Parents_i(\mathbf{X}))$ by the argument.\\
\\[-.5em]
In the reminder of this paper we will use $X_i$ to denote both the random variable and the corresponding node $i \in \mathbf{V}$.

\paragraph{Identifiability assumptions} In order to ensure identifiability of the causal graph from observational data, such that knowing the joint distributions of $\mathbf{X}$ is enough to distinguish causes and effects in the underlying causal graph, we need to make additional assumptions on the functional mechanisms $f_i$ of the ANM and on the distribution of the noise terms. In what follows, we provide identifiability conditions for a bivariate graph as in \cite{peters_2014_identifiability}. These can be seamlessly generalized to the multivariate case, which is discussed in Appendix \ref{app:multivariate_anm}. 
\begin{condition}[Condition 19 of \cite{peters_2014_identifiability}]\label{cond:identifiab}
    Given a  bivariate model $X_i \coloneqq N_i$ and $X_j \coloneqq f_j(X_i) + N_j$ with $\{i, j\} = \{1, 2\}$ generated according to \eqref{eq:scm}, we call the SEM an identifiable bivariate ANM if the triple $(f_j, p_{N_i}, p_{N_j})$ does not solve the following differential equation for all pairs $x_i, x_j$ with $f'_j(x_i)g''(x_j - f_j(x_i)) \neq 0$:
    \begin{equation}\label{eq:condition}
        k''' = k''\left(-\frac{g'''f'}{g''} + \frac{f''}{f'} \right) - 2g''f''f' + g'f''' + \frac{g'g'''f''f'}{g''} - \frac{g'(f'')^2}{f'} \:.
    \end{equation}
    Here, $f \coloneqq f_j$, $k \coloneqq \log p_{N_i}$, $g \coloneqq \log p_{N_j}$. The arguments $x_j - f_j(x_i)$, $x_i$ and $x_i$ of $g$, $k$ and $f$ respectively, have been removed to improve readability.
\end{condition}
 \cite{hoyer2009ANM} is the first to prove that if Condition \ref{cond:identifiab} is satisfied, then the graph associated with the bivariate ANM is identifiable from the joint distribution $p_{\mathbf{X}}$ (Theorem 20 \cite{peters_2014_identifiability}). 

Intuitively, we expect that Condition \ref{cond:identifiab} is satisfied for \textit{generic} triples $(f_j, p_{N_i}, p_{N_j})$. More formally, this is true because, for a fixed pair $(f_j, p_{N_j})$, the space of continuous distributions $p_{N_i}$ such that Condition \ref{cond:identifiab} is violated is contained in a three dimensional space. Since the space of continuous distributions is infinite dimensional, we can say that Condition \ref{cond:identifiab} is satisfied for "most" choices of $p_{N_i}$. For a rigorous statement see Proposition 21 in \cite{peters_2014_identifiability}. \\[.4em]
In practice, identifiability is satisfied if the noise terms have strictly positive densities $p_{N_i}, p_{N_j}$ and 
\begin{equation}\label{eq:f'g''}
    f'_j(x_i)g''(x_j - f_j(x_i)) \neq 0
\end{equation}
for all but a finite subset of points $(x_i, x_j)$ (\cite{zhang2009PNL}, Proposition 23 \cite{peters_2014_identifiability}). Additionally, we explicit that  Condition \ref{cond:identifiab} implies that 
\begin{equation}\label{eq:non_constant_score}
    \partial_{x_i} (k'(x_i) - f'_j(x_i)g'(x_j - f_j(x_i)) \neq 0 \:,
\end{equation}
\looseness-1for all $x_i, x_j$ such that $f'_j(x_i)g''(x_j - f_j(x_i)) \neq 0$. This can be directly verified in the proof of Theorem 20 (Equation 14) in \cite{peters_2014_identifiability}: if left hand side of \eqref{eq:non_constant_score} is null, then Equation \eqref{eq:condition} is always satisfied for $x_i, x_j$ with $f'_j(x_i)g''(x_j - f_j(x_i)) \neq 0$, which contradicts the conditions for identifiability. \\[.4em]
In the reminder of the work we consider these requirements (and their counterpart in the multivariate case as discussed in Appendix \ref{app:multivariate_anm}) to be satisfied by assumption in the SCM \eqref{eq:scm}.

\paragraph{Topological ordering definition}
\looseness=-1Let $\mathcal{G} = (\mathbf{X}, \mathcal{E})$ be a DAG. An ordering of the nodes $\mathbf{X}^{\pi} = X_{\pi_1}, \ldots, X_{\pi_d}$ is a topological ordering relative to $\mathcal{G}$ if, whenever we have $X_{\pi_i} \rightarrow X_{\pi_j} \in \mathcal{E}$, then $i < j$ (\cite{koeller_daphne_friedman}).

\paragraph{Problem definition}
Given an i.i.d. sample from the joint distribution $p_{\mathbf{X}}$, we want to infer the true causal graph $\mathcal{G}$ underlying the model that generated the data. One approach to solve this problem is to divide the task in two steps: first we find a topological ordering for the vertices in the graph, and then we prune the fully connected graph obtained drawing an edge from each node to all its successors in the ordering. In this paper, we use the score function $\nabla \log p_\mathbf{X}(\mathbf{X})$ to propose a consistent method of inference of the topological ordering  that works without assuming any specific distribution of the noise terms $N_i$ in Equation \eqref{eq:scm}.\\
\\[-.4em]
In practice, classical and state of the art causal discovery algorithms like SCORE (\cite{rolland_2022}), CAM (\cite{B_hlmann_2014}), GES (\cite{chickering_2003}) and GraN-DAG (\cite{grandag_19}) assume the noise terms to be normally distributed. In the next section we show the limitations of this assumption, and how the topological ordering can be wrongly inferred when it is violated, leading to the estimation of a graph with inverted edges.

\section{Limitations of the Gaussian noise assumption}\label{sec:gauss_hp}
It has been shown that for data generated by nonlinear models and additive noise, generally speaking, the causal direction between two variables is identifiable because in the reverse direction, one cannot find an independent residual (\cite{hoyer2009ANM,zhang2009PNL}), and hence the data likelihood given by the regression model (which assumes independent residuals) in the reverse direction is lower than that in the causal direction.  The identifiability results (\cite{zhang2009PNL}) imply that if the noise term in the causal model is Gaussian while the function is nonlinear, causal direction between two variables is identifiable. That is, the likelihood of the regression model in the correct direction is higher than that in the reverse direction, or equivalently, the total entropy of the estimated noise terms (including the hypothetical cause variable) is smaller in the causal direction (\cite{Zhang09_additive}).  

\vspace{0.7em}
However, this result does not imply that when the noise distribution is assumed to be Gaussian, the correct causal direction can always give a higher likelihood or lower total entropy of the estimated noise terms.

\begin{figure}
     \centering
     \begin{subfigure}[t]{0.24\textwidth}
         \centering
         \includegraphics[width=\textwidth]{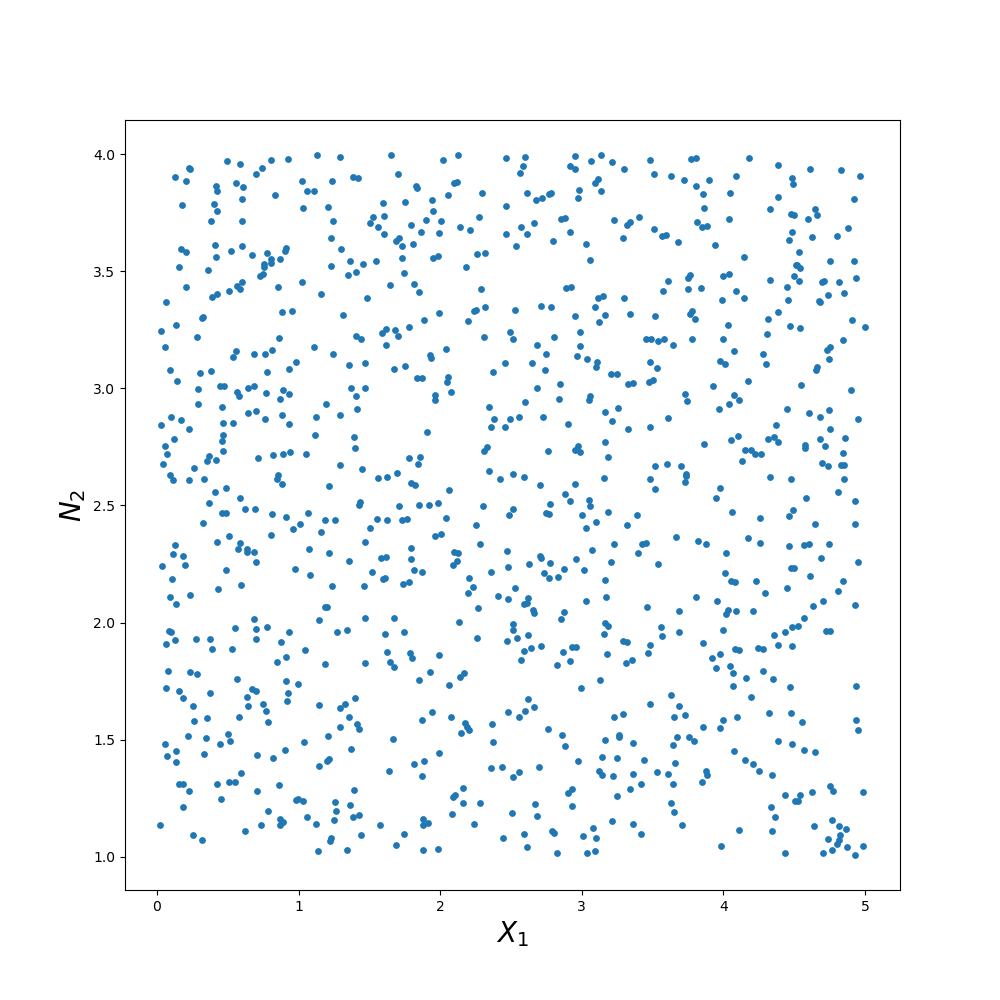}
         \caption{Scatter plot of $X_1$ and $N_2$}
         \label{fig:x1_eps2}
     \end{subfigure}
     \hfill
     \begin{subfigure}[t]{0.24\textwidth}
         \centering
         \includegraphics[width=\textwidth]{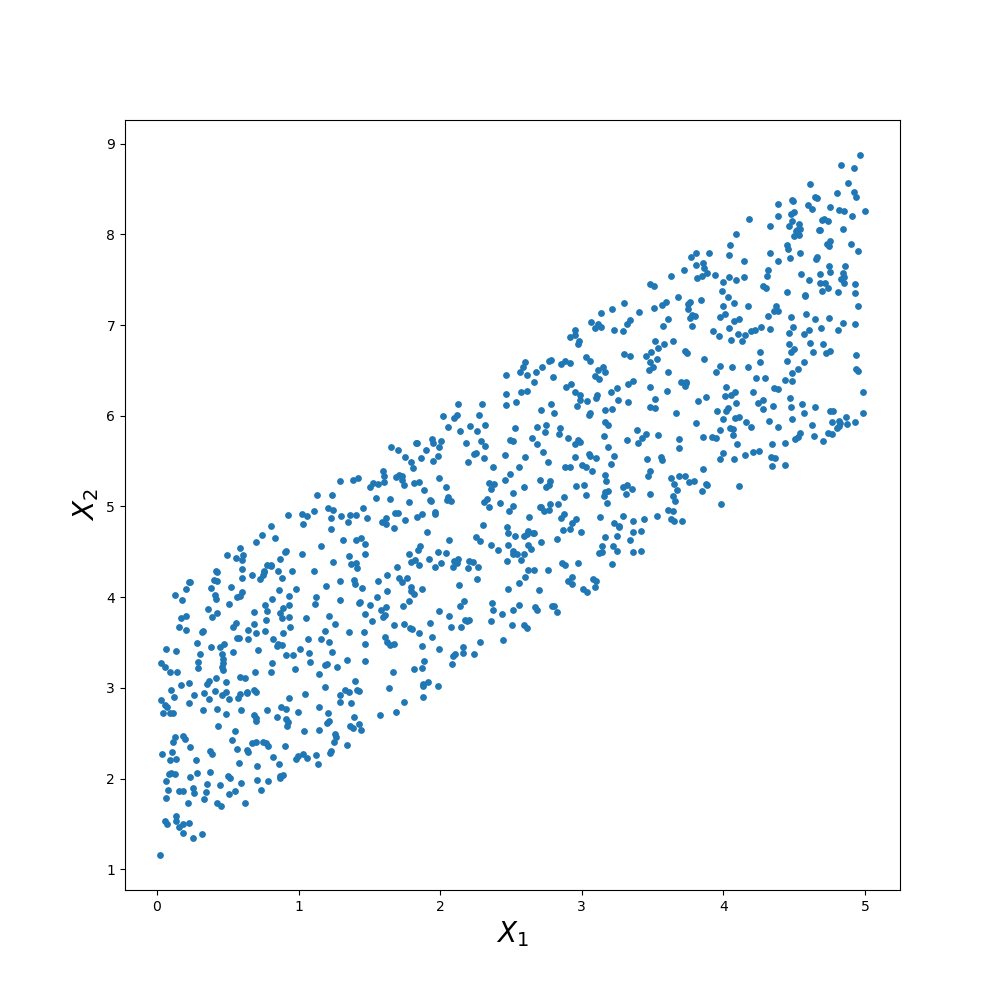}
         \caption{Scatter plot of $X_1$ and $X_2$}
         \label{fig:x1_x2}
     \end{subfigure}
     \hfill
     \begin{subfigure}[t]{0.24\textwidth}
         \centering
         \includegraphics[width=\textwidth]{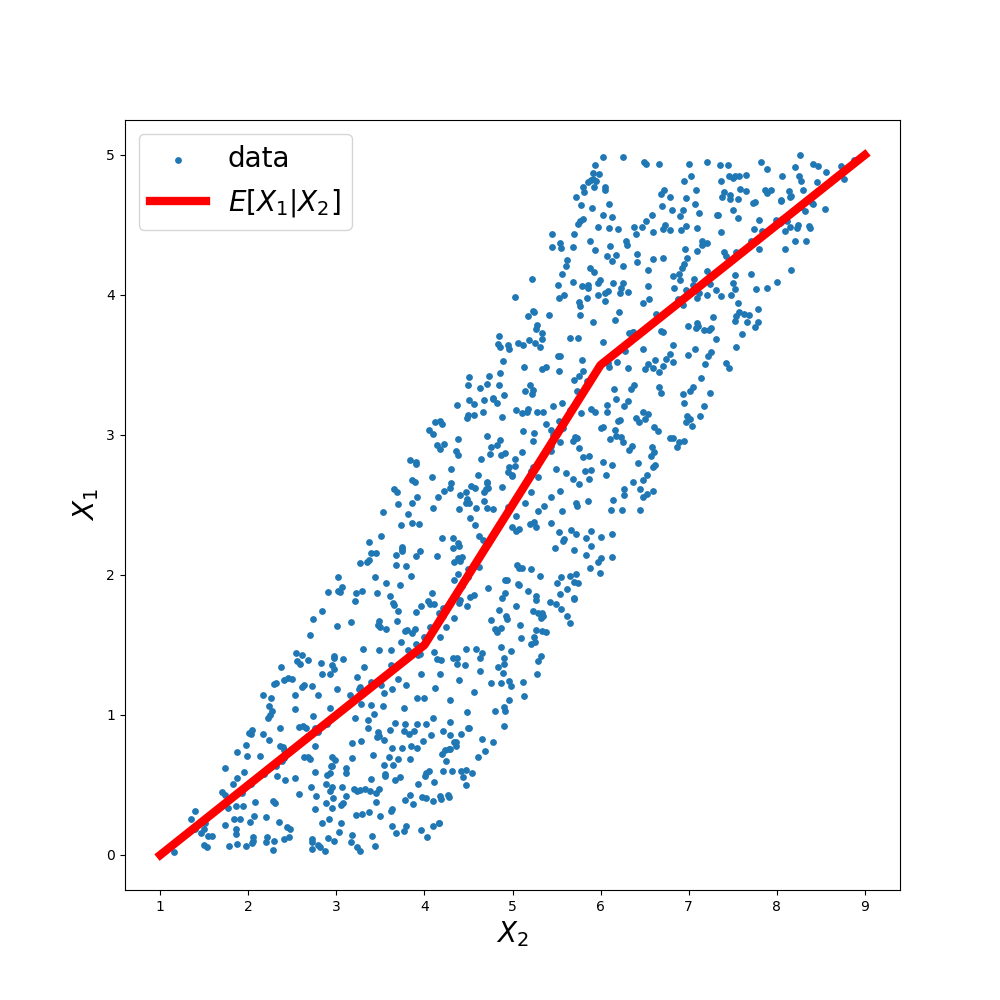}
         \caption{Scatter plot of $X_2$ and $X_1$ and regression curve}
         \label{fig:x2_x1}
     \end{subfigure}
          \hfill
     \begin{subfigure}[t]{0.24\textwidth}
         \centering
         \includegraphics[width=\textwidth]{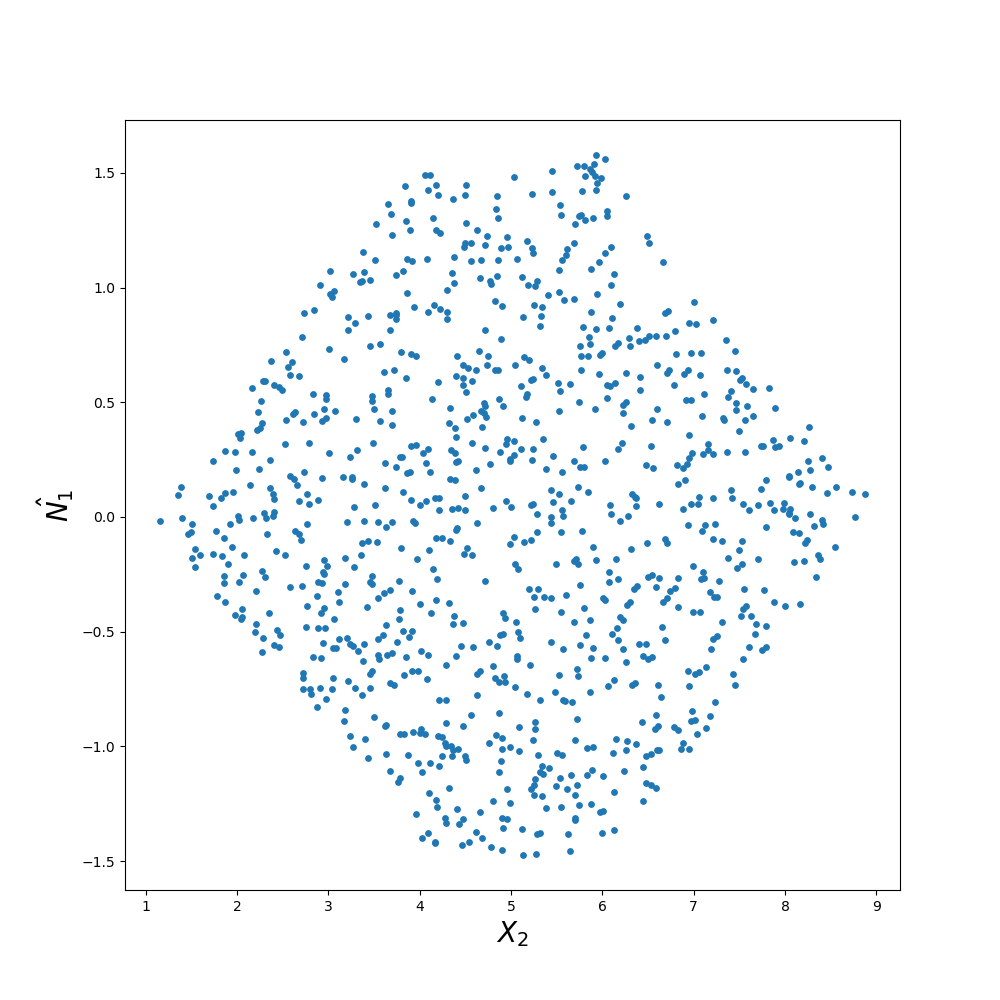}
         \caption{Scatter plot of $X_2$ and estimated $N_1$}
         \label{fig:x2_eps1}
     \end{subfigure}
        \caption{{Example to illustrate the limitations of the Gaussian noise assumption on the graph $X_1 \rightarrow X_2$.}}
        \label{fig:gauss_inversion}
\end{figure}

\begin{proposition}
\label{prop:gauss_violation}
Given a ground truth identifiable nonlinear additive causal model, any inference algorithm based on observational data that wrongly assumes Gaussianity of the noise terms is not guaranteed to recover the correct direction of the edges in the underlying graph.
\end{proposition}
In the reminder of the paper we consider Proposition \ref{prop:gauss_violation} to be true, and justify our claim with an example. We generate data from a bivariate model with additive uniform noise, and perform inference of the causal effects by comparing the total entropy of the residuals, as proposed in \cite{Zhang09_additive}. We show that assuming Gaussianity of the noise terms leads the method to failure, causing inference in the reversed direction. Additional details on the experimental design of the example can be found in Appendix \ref{app:example}. 

\begin{example}\label{inversion_example}
We start defining an example for a linear additive noise model, such that closed form solutions for the regression problems at hand can be found. Then, we generalize the example to the nonlinear case. 

\vspace{.7em}
Let $X_1 \rightarrow X_2$ with $X_2 \coloneqq X_1 + N_2$, where $X_1$ and $N_2$ are both uniformly distributed, as shown in Figure~\ref{fig:x1_eps2}. Figure~\ref{fig:x1_x2} shows the scatter plot of $X_1$ and $X_2$. Figure~\ref{fig:x2_x1} gives the scatter plot of $X_2$ and $X_1$, together with the regression curve, which, in this example, is piecewise linear.  Figure~\ref{fig:x2_eps1} shows the scatter plot of $X_2$ and $\hat{N}_1$, the estimated regression noise in the reverse direction. \\
One can find the correct causal direction by comparing the total entropy of the estimated noise terms. Let $H(\cdot)$ denote the differential entropy. One can calculate that 
$$H(X_1) + H(N_2) = 2.708 < H(X_2) + H(\hat{N}_1) = 2.954$$ where $H(X_1), H(X_2)$ are calculated exploiting knowledge of the marginal distributions.  $H(\hat{N}_1)$ is estimated from the data, while we use $H(N_2)$ in place of $H(\hat{N}_2)$ estimate as using the exact entropy makes computations more precise.
Let $H_\mathcal{G}(\cdot)$ denote the differential entropy under the Gaussianity assumption. One can then calculate that 
$$H_\mathcal{G}(X_1) + H_\mathcal{G}(N_2) = 3.061 > H_\mathcal{G}(X_2) + H_\mathcal{G}(\hat{N}_1) = 3.043.$$ 
That is, under the Gaussianity assumption, the reverse direction gives a lower total entropy (or equivalently, a higher likelihood), and hence a wrong causal direction is inferred.\\
Why does it happen? Notice that for a variable with a fixed variance, the Gaussian distribution gives the highest differential entropy. So naturally, $H_\mathcal{G}(X_1) + H_\mathcal{G}(N_2) > H(X_1) + H(N_2)$ and $H_\mathcal{G}(X_2) + H_\mathcal{G}(\hat{N}_1) > H(X_2) + H(\hat{N}_1)$. Furthermore, it is totally possible (which is clearly the case in this example) that compared to the original independent variables $X_1$ and $N_2$, $X_2$ and $\hat{N}_1$ are respectively closer to Gaussian. That is, the change induced by the Gaussianity assumption in the total entropy of the estimated noise terms is smaller in the reverse direction. As a consequence, under the Gaussianity assumption, the reverse direction may give lower total entropy of the estimated noise terms, in contrast to the case using their true distributions.

\vspace{0.7em}
Now consider the nonlinear model $X_2 \coloneqq X_1^{1 + \delta} + N_2$, with $\delta > 0$ and noise terms $X_1$ and $N_2$ uniformly distributed. Clearly, for $\delta = 0$ this is equivalent to the linear example already discussed: here we set $\delta = 0.1$ to introduce a weak nonlinearity in the generative process. Again, the ground truth
causal direction can be identified by comparing the total entropy of the estimated noise terms in the correct and reversed direction. One can calculate that $$H(X_1) + H(N_2) = 2.708 < H(X_2) + H(\hat{N}_1) = 2.926 \:.$$
Similarly, under Gaussianity assumption one obtains $$H_\mathcal{G}(X_1) + H_\mathcal{G}(N_2) = 3.061 > H_\mathcal{G}(X_2) + H_\mathcal{G}(\hat{N}_1) = 3.001 \:,$$ showing that the inversion statement of Proposition \ref{prop:gauss_violation} holds.
\end{example}

Given the shortcomings of the Gaussian assumption, we now propose a causal discovery method on additive nonlinear models with generic noise terms, such that the inferred causal ordering (and hence, the edges direction) is guaranteed to be correct with respect to the causal graph.

\section{Causal discovery via the score function}\label{sec:method}
In this section we derive a principled approach to identify leaf nodes from the score function $s(\mathbf{X}) = \nabla \log p_\mathbf{X}(\mathbf{X})$, without assuming any distribution of the noise random variables in the generative model \eqref{eq:scm} of $\mathbf{X}$.

\subsection{Score function of a data distribution}
Given the distribution $p_\mathbf{X}(\mathbf{X})$ induced by model \eqref{eq:scm}, we can define the vector of the score function as $s(\mathbf{X}) = \nabla \log p_\mathbf{X}(\mathbf{X})$. Exploiting the factorization of the joint distribution of Equation \eqref{eq:factorization}, we can write:
\begin{equation}
    \log p_\mathbf{X}(\mathbf{X}) = \sum_{i=1}^d \log p_i(X_i \mid \Parents_i) \:.
\end{equation}
such that a single entry $s_i(\mathbf{X})$ of the score is equal to:
\begin{equation}
    s_i(\mathbf{X}) = \partial_{\mathsmaller{X_i}} \log p_i(X_i \mid \Parents_i) + \sum_{k \in \Child_i} \partial_{\mathsmaller{X_i}}f_k(\Parents_k)\partial_{\mathsmaller{X_k}} \log p_k(X_k \mid \Parents_k) \; .
    \label{eq:score_i}
\end{equation}
Under parents conditioning, the marginal of $X_i$ is the same as the distribution of $N_i$ shifted by the value of the mechanism $f_i(\Parents_i)$, meaning that $p_i(X_i\mid \Parents_i)$ can be replaced by $p_i(N_i = X_i - f_i(\Parents_i) \mid  \Parents_i)$. This allows to rewrite the score as:
\begin{equation}
        s_i(\mathbf{X}) = \partial_{\mathsmaller{N_i}} \log p_i(N_i) - \sum_{k \in \Child_i} \partial_{\mathsmaller{X_i}}f_k(\Parents_k)\partial_{\mathsmaller{N_k}} \log p_k(N_k) \; .
\end{equation}
\looseness-1Then, for each $i = 1, \ldots, d$ we define a function $g_i(N_i) \coloneqq \log p_i(N_i)$, such that the $i$-th score entry is
\begin{equation}
        s_i(\mathbf{X}) = \partial_{\mathsmaller{N_i}} g_i(N_i) - \sum_{k \in \Child_i} \partial_{\mathsmaller{X_i}}f_k(\Parents_k)\partial_{\mathsmaller{N_k}} g_k(N_k) \; .
        \label{eq:score_g}
\end{equation}
For a leaf node $X_l$, Equation \eqref{eq:score_g} of the score becomes 
\begin{equation}
    s_l(\mathbf{X}) = \partial_{\mathsmaller{N_l}} g_l(N_l) \:,
    \label{eq:score_nl}
\end{equation}
telling us that, if $g_l$ were known, we could predict the score of a leaf $s_l(\mathbf{X})$ from the noise $N_l$. 
In the next section, with this idea in mind, we define a regression problem for each variable $X_i$, where the input variables are all the remaining entries $\Xminus{i} \coloneqq \mathbf{X} \setminus \{X_i\}$: then we show that if $X_i$ target of the prediction is a leaf, the residuals of this learning problem are consistent estimators of the noise term in the corresponding structural equation of model \eqref{eq:scm}. 

\subsection{Residuals estimation}
Given an i.i.d. sample $X \in \R^{n \times d}$ from $p_\mathbf{X}$, for each $i = 1, \ldots, d$ we define a regression problem predicting $X_i$ from the remaining variables $\Xminus{i}$:
\begin{equation}
    \begin{split}
        & \min_{q \in \mathcal{Q}} L(q), \;\; L(q) = \int_{\R^d} (q(\Xminus{i}) - X_i)^2 dp_\mathbf{X}(\mathbf{X}) \\
        & \textnormal{given} \; \mathcal{D} = \left\{\left(\Xminus{i}^k, X_i^k\right)\right\}_{k=1}^n
    \end{split} \:\:\:\:,
\label{eq:res_problem}
\end{equation}
where $\mathcal{Q}$ is the space of measurable functions from the input space to $\R$.
For all $\Xminus{i}$ in the input space, the minimizer $q^* \in \mathcal{Q}$ of $L$ is the conditional expectation $\operatorname{\mathbf{E}}[X_i \mid \Xminus{i}]$, which by linearity of the mean is equivalent to:
\begin{equation}
    q^*(\Xminus{i})
    = \operatorname{\mathbf{E}}[f_i(\Parents_i) \mid \Xminus{i}]  + \operatorname{\mathbf{E}}[N_i \mid \Xminus{i}]\,.
\end{equation}
Given that $\Parents_i \subset \Xminus{i}$, we can simply remove the expectation operator from the first term of the sum, obtaining: 
\begin{equation}
    q^*(\Xminus{i})
    = f_i(\Parents_i) + \operatorname{\mathbf{E}}[N_i \mid \Xminus{i}]\,.
    \label{eq:q_star}
\end{equation}
Now we define the residual of the learning problem in \eqref{eq:res_problem} as the difference between the response and the target function:
\begin{equation}
\begin{split}
    R_i &\coloneqq X_i - q^*(\Xminus{i}) \\
    &= N_i - \operatorname{\mathbf{E}}[N_i \mid \Xminus{i}] \:,
\end{split}
\label{eq:residual}
\end{equation}
where the second equality holds from Equation \eqref{eq:q_star}.
We can further manipulate the residual expression by noticing that $\Xminus{i} = \Desc_i \cup \Nondesc_i$, with $\Desc_i$ and $\Nondesc_i$ respectively the set of descendants and non-descendants of a node $X_i$: Equation \eqref{eq:residual} then becomes
\begin{equation}
    R_i = N_i - \Mu\left[N_i \mid \Desc_i \cup \Nondesc_i \right] \:.
    \label{eq:res_leaf_exp}
\end{equation}
For a leaf node $X_l$ we can exploit the fact that $\Desc_l = \emptyset$ in order to simplify the above expression of the residual in $R_l = N_l - \Mu\left[N_l \mid \Nondesc_l \right]$. Additionally, we can use d-separation criterion to conclude that $N_l$ is unconditionally independent of $\Nondesc_l$, as shown in Figure  \ref{fig:d-sep}: the expectation on the error term of a leaf is then $\Mu\left[N_l \mid \Nondesc_l \right] = \Mu\left[N_l \right]$. Finally, under the assumption of zero mean of the noise of model \eqref{eq:scm}, we conclude that the residual of Equation \eqref{eq:res_leaf_exp} is
\begin{equation}
    R_l = N_l \:.
\end{equation}
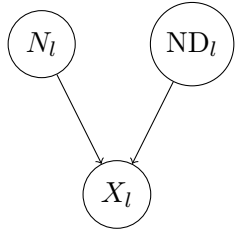
\begin{figure}
    \centering
    \begin{tikzpicture}
        \node[shape=circle,draw=black] (A) at (1,0) {$\Nondesc_l$};
        \node[shape=circle,draw=black] (B) at (-1,0) {$N_l$};
        \node[shape=circle,draw=black] (C) at (0,-2) {$X_l$};
        
        \draw[->] (A) -- (C);
        \draw[->] (B) -- (C);
    \end{tikzpicture}
    \caption{{Consider a leaf node $X_l$ generated according to the SCM defined in \eqref{eq:scm}: by \textit{d-separation} the noise term $N_l$ is unconditionally independent from the set of non-descendants $\Nondesc_l$, as the only path between the two nodes contains a collider, namely $X_l$.}}
    \label{fig:d-sep}
\end{figure}
Exploiting this equivalence we can rewrite the score of a leaf in Equation  \eqref{eq:score_nl} as a function of $R_l$, such that the score entry $s_l$ satisfies:
\begin{equation}
    \boxed{s_l(\mathbf{X}) = \partial_{\mathsmaller{N_l}} g_l(R_l)} \:.
    \label{eq:score_rl}
\end{equation}
By substituting the residual of Equation \eqref{eq:residual} in Equation \eqref{eq:score_g}, we can derive an analogous expression of the score entry $s_i$ associated to a non leaf node $X_i$:
\begin{equation}
    s_i(\mathbf{X}) = \partial_{\mathsmaller{N_i}} g_i(R_i + \operatorname{\mathbf{E}}\left[N_i \mid \Xminus{i} \right]) - \sum_{k \in \Child_i} \partial_{\mathsmaller{X_i}}f_k(\Parents_k)\partial_{\mathsmaller{N_k}} g_k(R_k + \operatorname{\mathbf{E}}\left[N_k \mid \Xminus{k} \right])) \; .
    \label{eq:score_i_generic}
\end{equation}

\paragraph{Discussion} 
For a leaf node $X_l$, Equation \eqref{eq:score_rl} shows that the associated score entry is a function of a single variable, namely the residual $R_l$: this suggests that we can hope to consistently estimate the score $s_l(\mathbf{X})$ from such residual. On the other hand, if we consider a non-leaf node $X_i$, the form associated to its score $s_i(\mathbf{X})$ is more complicated (i.e. depending on a larger number of variables), as shown by Equation \eqref{eq:score_i_generic}: intuitively we can see that $R_i$, as a predictor, is not sufficient to find a consistent estimator of $s_i(\mathbf{X})$. In the next section, we want to formalize these intuitions that will allow us to derive a theoretically principled method to identify leaf nodes by looking at the error of score entries predictions from their corresponding residual.

\subsection{Identifying leaf nodes}\label{sec:leaf_node}
Consider a leaf node $X_l$: given a set of i.i.d. observations $\left\{(R_l^k, s_l(\mathbf{X}^k))\right\}_{k=1}^n$, we want to find an estimator of the score using the residuals as input. Similarly to \eqref{eq:res_problem} we define a regression problem
\begin{equation}
        \min_{h \in \mathcal{H}} L(h), \;\; L(h) = \int_{\R \times \R^d} (h(R_l) - s_l(\mathbf{X}))^2 dp(R_l, \mathbf{X}) \:,
\label{eq:score_problem}
\end{equation}
with $\mathcal{H}$ set of measurable functions from input to output space.
For every $R_l \in \R$, the target function minimizing the expected risk is 
\begin{equation}
    \begin{split}
        h^*(R_l) &\coloneqq \Mu\left[s_l(\mathbf{X}) \mid R_l\right] \\
        &= \Mu\left[\partial_{\mathsmaller{N_l}} g_l(R_l) \mid R_l\right] \:.
    \end{split}
\label{eq:reg_score}
\end{equation}
It is immediate to see that, since we are conditioning on $R_l$, we can remove the expectation operator, obtaining:
\begin{equation}
        h^*(R_l) = \partial_{\mathsmaller{N_l}} g_l(R_l), \: \forall R_l \in \R \:.
\end{equation}
Then, given a sample $(R_l, s_l(\mathbf{X}))$, the difference between the prediction $h^*(R_l)$ and the ground truth $s_l(\mathbf{X}) = g_l(R_l)$ is
\begin{equation}
    \boxed{h^*(R_l) - s_l(\mathbf{X}) = 0} \:.
    \label{eq:zero_err}
\end{equation}
Similarly, the regression problem of \eqref{eq:score_problem} can be defined for a non-leaf node $X_i$: the resulting regression function is $h^*(R_i) = \operatorname{\mathbf{E}}\left[s_i(\mathbf{X}) \mid R_i \right]$, the conditional expectation of Equation  \eqref{eq:score_i_generic}. Now, we can exploit these results to define a criterion for identification of leaf nodes.

\begin{lemma}
\label{lem:lemma}
Let $\mathbf{X}$ be a random vector generated according to model \eqref{eq:scm}, and let $X_i \in \mathbf{X}$. Then $$\Mu\left[\left(h^*(R_i) - s_i(\mathbf{X})\right)^2\right] = 0 \Leftrightarrow X_i \textnormal{ is a leaf.}$$
\end{lemma}

\paragraph{Discussion} With Equation \eqref{eq:zero_err}, we show that we can find a consistent estimator $h^*$ that can exactly predict the score function associated to a leaf $X_l$, given that we observe the residual $R_l$. In general, this is not the case for a node $X_i$ that is not a leaf. This intuition is formalized in Lemma \ref{lem:lemma}, by considering the mean of the squared error of the prediction over all input and output realizations. The proof of the lemma can be found in Appendix \ref{ap:lem_proof}.

\vspace{.7em}
Based on the results of this section, we now introduce an algorithm for causal discovery that runs on nonlinear additive models with generic distributions of the noise terms. Then, we compare its experimental performance against several existing methods on synthetic data. 

\section{Method}\label{sec:algorithm}
In Section \ref{sec:leaf_node} we show how to identify leaf nodes in a causal graph underlying observations generated according to model \eqref{eq:scm}, consistently with the number of samples: the idea is that, given a set of observations, for each node $i = 1, \ldots, d$ we can predict the score $s_i(\mathbf{X})$ from the corresponding residual $R_i$, choosing as leaf the node $l$ index of the entry where the generalization error is minimized. Once a leaf is identified, it is removed from the graph, and the procedure is repeated iteratively up to the source node, allowing to infer a topological ordering $\mathbf{X}^\pi$ that is asymptotically consistent.
In practice, given a finite set of observations $X \in \R^{n \times d}$, first we estimate $\nabla \log p(\mathbf{X})$ score function of the data using the Stein gradient estimator (\cite{stein_gradient}, see Appendix \ref{app:score} for details), whose output is the vector $\hat{\mathbf{s}}(\mathbf{X})$. Then, we estimate the residuals $\hat{\mathbf{R}} = (\hat{R}_1, \ldots, \hat{R}_d)$ by Kernel Ridge regression, solving the problem defined in \eqref{eq:res_problem}. Next, we define the vector estimator $\tilde{\mathbf{s}}(\hat{\mathbf{R}})$ predicting $\hat{s}_i(\mathbf{X})$ from $\hat{R}_i$, for each variable $i = 1, \ldots, d$: in order to avoid overfitting, we train K different models by K-fold cross validation (i.e. only on a subset of the observations), each one predicting on its corresponding test set, unseen during the training. Finally we compute the Mean Squared Error (MSE) between the predictions $\tilde{\mathbf{s}}(\hat{\mathbf{R}})$ and the ground truth $\hat{\mathbf{s}}(\mathbf{X})$ provided by SCORE's output: we select as leaf the node corresponding to the \textit{argmin} of the vector of MSEs.  This procedure is repeated such that at each iteration one leaf is identified and added to the topological ordering estimate. More details on the implementation of the algorithm described can be found in the box of Algorithm \ref{alg:top_order}.

Given the order estimated by Algorithm \ref{alg:top_order}, we use a pruning method, namely the pruning procedure of CAM (\textit{CAM-pruning}, Appendix \ref{app:cam}), to remove superfluous edges from the fully connected graph admitted by the ordering.

\begin{algorithm}
\footnotesize
\setstretch{1.2}
    \caption{NoGAM causal discovery}
    \label{alg:top_order}
    
    Input: data matrix $X \in \mathbb{R}^{n \times d}$\;
    
    $X^\pi \leftarrow [\:]$\;
    
    $nodes \leftarrow [1, \ldots, d]$\;
    
    \For{ $i = 1, \ldots, d$ }{
    $\hat{\mathbf{s}} \leftarrow \textnormal{SCORE}(X)$\;
    
    $\hat{\mathbf{R}} \leftarrow \{\hat{X}_j\}_{j=1}$ estimate from $\Xminus{i}$\;
    
    $\tilde{\mathbf{s}} \leftarrow \{\tilde{s}_j\}_{j=1}$ estimate from $\hat{R}_i$\;
    
    $MSE \leftarrow \operatorname{Avg}\left[\hat{\mathbf{s}} - \mathbf{\tilde{s}}\right]^2$\;
    
    $l_{index} \leftarrow \operatorname{argmin} MSE$\;
    
    $l \leftarrow nodes[l_{index}]$\;
    
    $X^\pi \leftarrow \left[l, X^\pi\right]$\;
    
    Remove $l_{index}$-th column from $X$; Remove $l$ from $nodes$\;
    
    }
    
    $X^\pi \leftarrow reverse(X^\pi)$ (first node is source, last node is leaf)
    
    $\hat{\mathcal{G}} \leftarrow \textnormal{CAM-pruning}(X^\pi)$ (CAM-pruning: pruning method of CAM algorithm)
    
    \Return $\hat{\mathcal{G}}$
\end{algorithm}

\section{Experiments}\label{sec:experiments}
In this section we empirically study the performance of NoGAM (Algorithm \ref{alg:top_order}). The experimental analysis is done on data synthetically generated using the \cite{Erdos:1960} (ER) model. Mimicking the setting of \cite{rolland_2022}, \cite{grandag_19} and \cite{Zhu2020Causal}, we generate the mechanisms $f_i$ by sampling Gaussian processes with a unit bandwidth RBF kernel. Experiments are repeated with the number of nodes $d$ equals $10$ and $20$, and expected number of edges  equals to $d$ (ER1) and $4d$ (ER4), to simulate inference on sparser and denser graphs. The size of the datasets is $N=1000$. Performance is tested on datasets generated with noise terms under one of the following distributions: Beta, Exponential, Gamma, Gumbel, Laplace and Normal. Comparing the performance of NoGAM with state of the art methods working under Gaussianity assumption, we are able to provide empirical evidence of the robustness of our algorithm with respect to changes in the distributions.

The residuals $\hat{\mathbf{R}}$ and the the score function entries $\tilde{\mathbf{s}}(\hat{\mathbf{R}})$  are estimated by Kernel Ridge regression, using the \texttt{scikit-learn} (\cite{scikit-learn}) implementation of the algorithm, with hyperparameters $\alpha = 0.01$, $\gamma=0.1$: these values are tuned minimizing the generalization error on the estimated residuals, without using the performance on the causal graph ground truth. For the CAM pruning step the cutoff threshold is set at 0.001.

The metrics used to assess the quality of the inferred graph are the Structural Hamming Distance (SHD), which is the sum of false positive, false negative and reversed edges, and Structural Interventional Distance (SID, \cite{sid_2013}), accounting for the number of miscalculated interventional distributions from the inferred graph. We separately evaluate the quality of the topological ordering estimation as follow: given an ordering $\hat{\pi}$ and the ground truth adjacency matrix $A$, we use the topological ordering divergence defined in SCORE (\cite{rolland_2022}):
\begin{equation}
    D_{top}(\hat{\pi}, A) = \sum_{i=1}^d \sum_{j: \hat{\pi}_i \succ \hat{\pi}_j} A_{ij} \:, 
\end{equation}
with $\hat{\pi}_i \succ \hat{\pi}_j$ meaning the node $i$ is successive to $j$ in the order, and $A_{ij}=1$ if $X_i \in \Parents_j(\mathbf{X})$. In words this is the sum of the edges that can not be recovered due to the choice of the topological ordering. If $\hat{\pi}$ is correct with respect to $A$, then $D_{top}(\hat{\pi}, A) = 0$.

\vspace{0.7em}
We compare the experimental performance of our algorithm against SCORE (\cite{rolland_2022}), CAM (\cite{B_hlmann_2014}) and GES (\cite{chickering_2003}), causal discovery methods working under the assumption of Gaussian noise. We exclude PC and FCI since in general they perform much worse (\cite{B_hlmann_2014}, \cite{grandag_19}).

\begin{figure}
     \centering
     \begin{subfigure}[b]{0.32\textwidth}
         \centering
         \includegraphics[width=\textwidth]{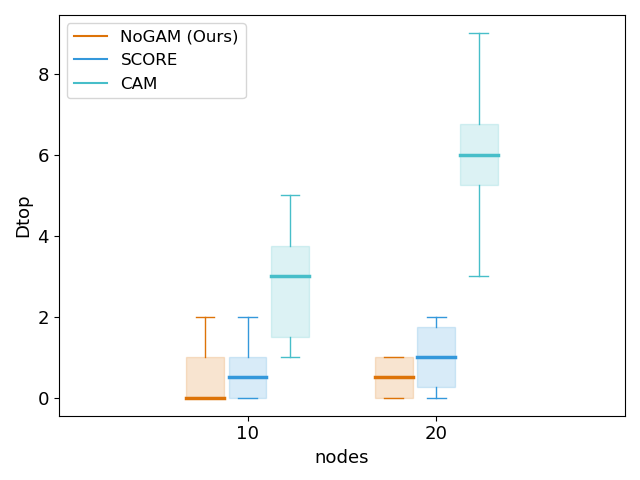}
         \caption{Beta distribution}
     \end{subfigure}%
     \hfill
     \begin{subfigure}[b]{0.32\textwidth}
         \centering
         \includegraphics[width=\textwidth]{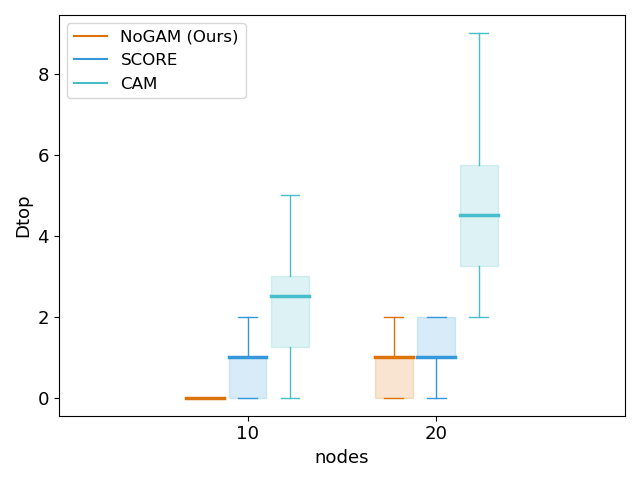}
         \caption{Exponential distribution}
     \end{subfigure}%
     \hfill
     \begin{subfigure}[b]{0.32\textwidth}
         \centering
         \includegraphics[width=\textwidth]{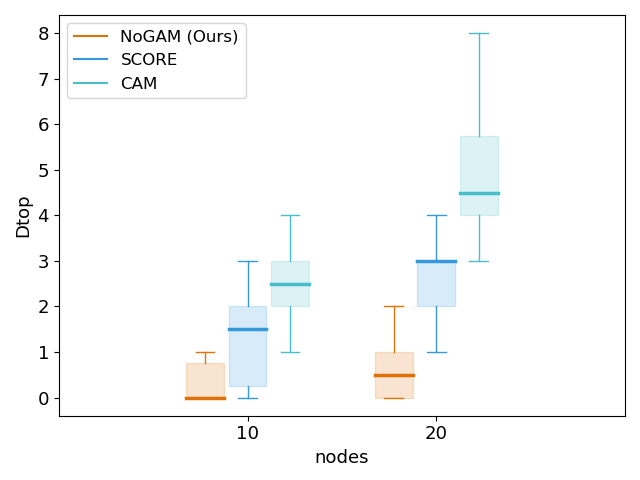}
         \caption{Gamma distribution}
     \end{subfigure}%
     
     \medskip
          \begin{subfigure}[b]{0.32\textwidth}
         \centering
         \includegraphics[width=\textwidth]{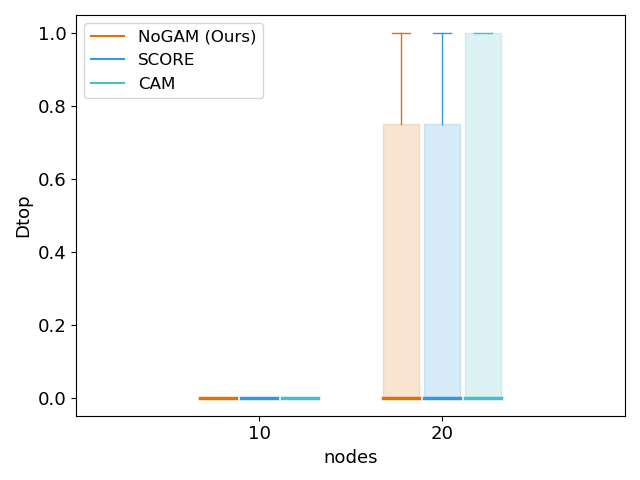}
         \caption{Normal distribution}
     \end{subfigure}%
     \hfill
     \begin{subfigure}[b]{0.32\textwidth}
         \centering
         \includegraphics[width=\textwidth]{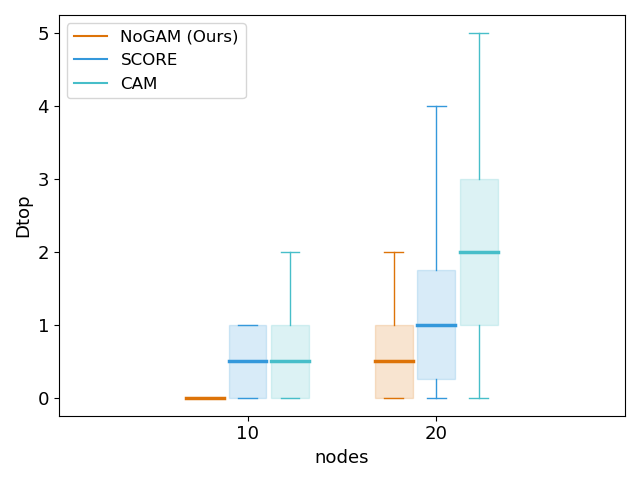}
         \caption{Gumbel distribution}
     \end{subfigure}%
     \hfill
     \begin{subfigure}[b]{0.32\textwidth}
         \centering
         \includegraphics[width=\textwidth]{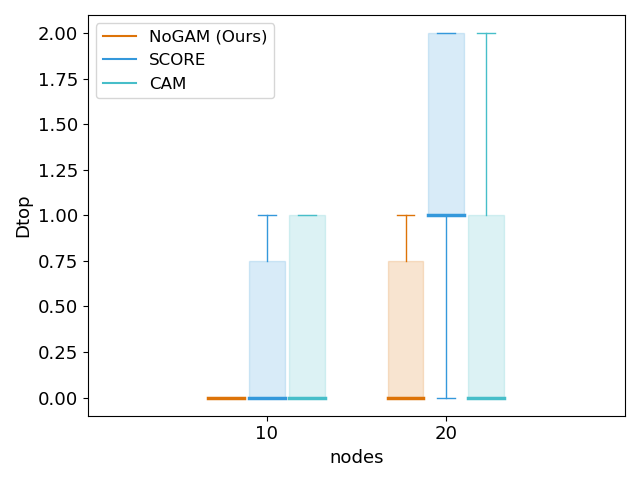}
         \caption{Laplace distribution}
     \end{subfigure}
        \caption{{Boxplots over 10 runs showing topological ordering divergence $D_{top}$ (the lower the better) over sparse graphs ER1. GES algorithm does not appear since it does not require an explicit estimate of the topological ordering. Overall, NoGAM clearly outperforms CAM and SCORE.}} 
        \label{fig:dtop_boxplots_er1}
\end{figure}

\begin{figure}
     \centering
     \begin{subfigure}[b]{0.32\textwidth}
         \centering
         \includegraphics[width=\textwidth]{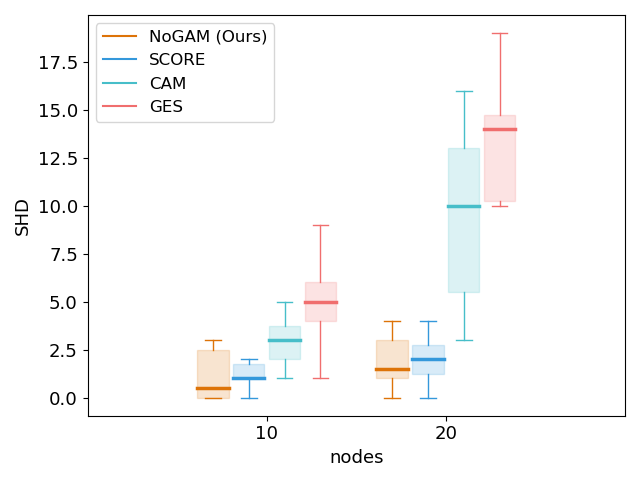}
         \caption{Beta distribution}
     \end{subfigure}%
     \hfill
     \begin{subfigure}[b]{0.32\textwidth}
         \centering
         \includegraphics[width=\textwidth]{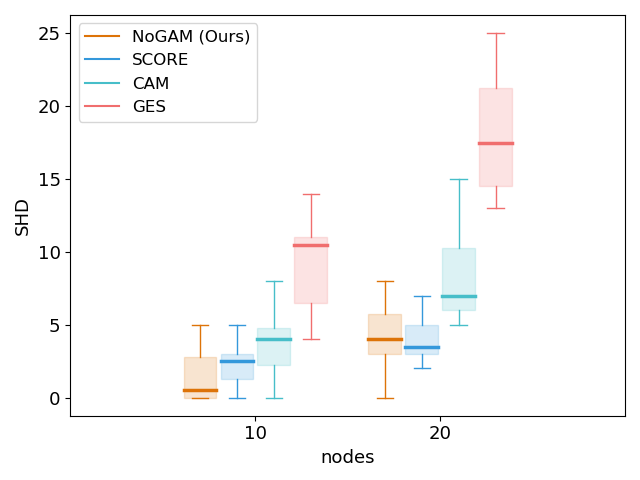}
         \caption{Exponential distribution}
     \end{subfigure}%
     \hfill
     \begin{subfigure}[b]{0.32\textwidth}
         \centering
         \includegraphics[width=\textwidth]{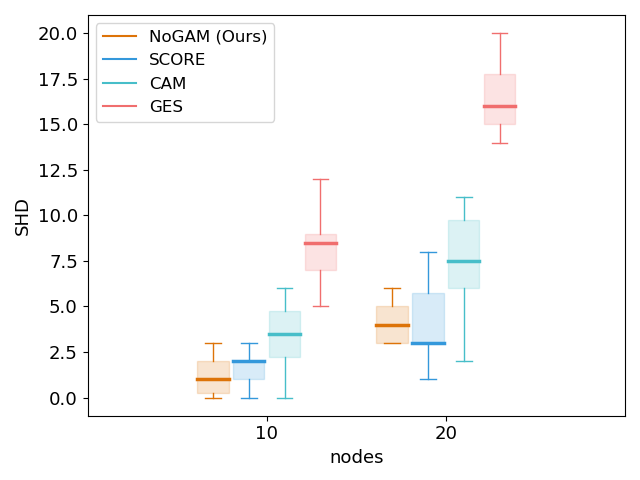}
         \caption{Gamma distribution}
     \end{subfigure}%
     
     \medskip
          \begin{subfigure}[b]{0.32\textwidth}
         \centering
         \includegraphics[width=\textwidth]{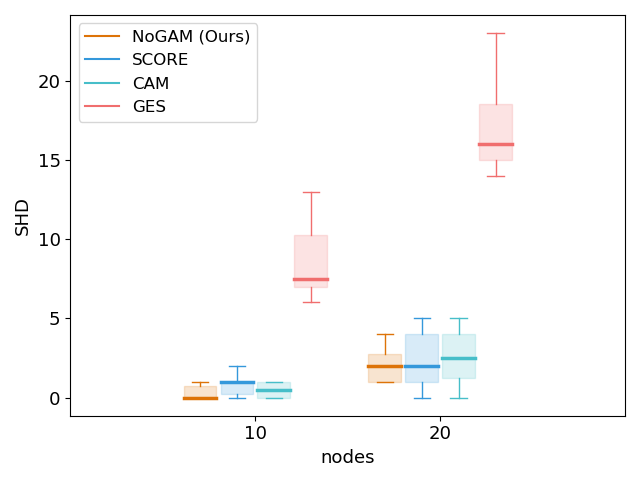}
         \caption{Normal distribution}
     \end{subfigure}%
     \hfill
     \begin{subfigure}[b]{0.32\textwidth}
         \centering
         \includegraphics[width=\textwidth]{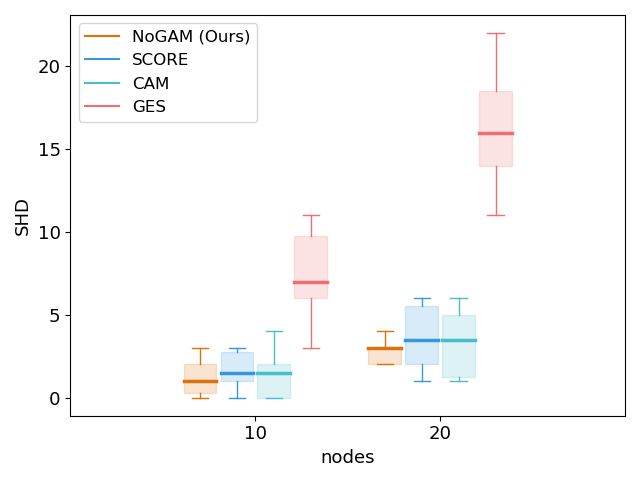}
         \caption{Gumbel distribution}
     \end{subfigure}%
     \hfill
     \begin{subfigure}[b]{0.32\textwidth}
         \centering
         \includegraphics[width=\textwidth]{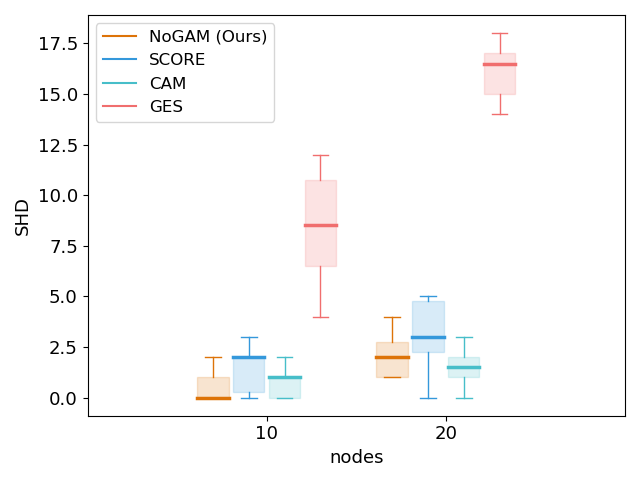}
         \caption{Laplace distribution}
     \end{subfigure}
        \caption{{Boxplots over 10 runs showing SHD performance over sparse graphs ER1. NoGAM is in general better than all the other methods, with SCORE showing comparable performance.}} 
        \label{fig:shd_er1}
\end{figure}

\vspace{.7em}
\looseness-1Figure \ref{fig:dtop_boxplots_er1} illustrate how in the sparse setting (ER1) topological ordering estimates of NoGAM systematically outperform the $D_{top}$ results obtained by CAM and SCORE for every non-Gaussian noise distribution. The accuracy gap closes only for datasets generated under Gaussianity of the noise, coherently with our expectations. A similar performance gap in favor of our method is observed in the dense case (ER4), as shown in Figure \ref{fig:dtop_boxplots_er4} of the Appendix. Overall SCORE and NoGAM clearly show better SHD (Figure \ref{fig:shd_er1} and \ref{fig:shd_er4}) with respect to the remaining methods. Note that the two algorithms differ only for the topological order inference step, while they share the same pruning method of CAM. A complete overview of the experimental results on ER graphs can be found in the  Appendix \ref{app:er}.\\[-.7em]
In the Appendix we provide a significant extension on our experiments. NoGAM performance is tested on Sachs real data (\cite{sachs_2005}, Appendix \ref{app:sachs}) and Scale-free synthetic graphs (\cite{Barabasi99emergenceScaling}, Appendix \ref{app:sf}). In Appendix \ref{app:hp_restriction} we analyze the performance of NoGAM under restriction of the hypothesis space to linear functions for the regression problems of Equations \eqref{eq:res_problem} and \eqref{eq:reg_score}. Additionally, in Appendix \ref{app:comparison_mooij_peters} we compare with  \cite{mooij2009} and \cite{peters_2014_identifiability}, which proposal rely on independence of the residuals to discover causal effects on nonlinear ANM without restrictions on the noise terms distribution. 


\begin{paragraph}{Discussion on SCORE robustness}
\looseness-1Despite being systematically outperformed by our method (Figure \ref{fig:dtop_boxplots_er1} and \ref{fig:dtop_boxplots_er4}), SCORE algorithm shows significant robustness across different distributions of the noise terms. Here we want to provide a brief discussion on why this, in our opinion, is the case. In Lemma 1 of \cite{rolland_2022}, authors propose to identify a leaf node from the Jacobian matrix of the score function $\mathbf{s}(\mathbf{X})$. In particular, computing the variance of the diagonal elements of the Jacobian $\operatorname{Var}\left[\partial_{\mathsmaller{X_i}} s_i(\mathbf{X}) \right], \: \forall i = 1, \ldots, d$, it can be shown the following: under Gaussianity assumption, leaf nodes are associated to zero variance, such that $\operatorname{Var}\left[\partial_{\mathsmaller{X_l}} s_l(\mathbf{X}) \right] = 0 \Longleftrightarrow \Child_l(\mathbf{X}) = \emptyset$, i.e. if and only if $X_l$ is a leaf. In practice, due to statistical error in the estimation, such expression in general is never exactly zero. To account for this, the SCORE algorithm iteratively selects leaf nodes as $l = \operatorname{argmin}_{i \in \{1, \ldots, d\}} \operatorname{Var}\left[\partial_{\mathsmaller{X_i}} s_i(\mathbf{X}) \right]$. We argue that such heuristic is key to determine the robustness of the algorithm as empirically observed outside of the Gaussian assumption. If the noise terms are not normally distributed, then the score function is equivalent to Equation \eqref{eq:score_i}.
\looseness-1It is clear that the variance of $\partial_{X_i}s_i(\mathbf{X})$ is proportional to the number of children in the summation term. In particular, the total variance of a diagonal element of the Jacobian of the score is the sum of the variances of the partial derivative of the two right-hand terms in Equation \eqref{eq:score_i}, plus their covariance: if the covariance happens to be negative and with magnitude large enough, the variance associated to a non-leaf node might be smaller than the one relative to all other nodes, leading to errors in the inferred topological ordering. Nevertheless, in general, this doesn't seem to be the case:  despite the guarantee of vanishing variance for leaf nodes doesn't hold anymore, in practice the leaf selection criterion based on the $\operatorname{argmin}$ operator can still be expected to work. 
In Appendix \ref{app:score} we extend the discussion on \cite{rolland_2022}, focusing on the differences of our method. 
\end{paragraph}

\subsection{Algorithmic complexity}
We now provide an analysis of the algorithmic complexity of NoGAM topological ordering method. We denote with $n$ the number of samples and $d$ the number of nodes. Considering the implementation of Algorithm \ref{alg:top_order}, each iteration of the for loop needs to solve a regression problem with $\mathcal{O}(n^3)$ cost, for each of the $d$ residuals $\hat{R}_i$ estimated. The same analysis holds for the estimation of $\tilde{\mathbf{s}}$ from $\hat{\mathbf{R}}$. This provides an overall $\mathcal{O}(d^2n^3)$ complexity.
Similarly to NoGAM, SCORE iteratively identifies leaves one at the time: each iteration requires inverting the $n \times d$ dimensional matrix of the data, such that the topological ordering inference time scales with $\mathcal{O}(dn^3)$.
An overview of the execution times of the experiments is provided in Table \ref{tab:exe_time}.
\begin{table}[]
\footnotesize
    \centering
    \begin{tabular}{lccc}
        \midrule
         & Method & Ordering time [s] & Total time [s] \\
         \midrule
         \multirow{4}{*}{d=10} & NoGAM & $6.98 \pm 0.57$ & $10.56 \pm 0.84$ \\
         & SCORE & $4.28 \pm 0.28$ & $7.72 \pm 01.07$ \\
         & CAM & $24.12 \pm 1.58$ & $27.98 \pm 2.61$  \\
         & GES & $-$ & $0.43 \pm 0.11$ \\
         \midrule
         \multirow{4}{*}{d=20} & NoGAM & $25.75 \pm 1.48$ & $53.98 \pm 2.01$ \\
         & SCORE & $9.51 \pm 1.12$ & $37.3 \pm 4.21$ \\
         & CAM & $501.92 \pm 11.92$ & $529.61 \pm 18.41$   \\
         & GES & $-$ & $5.35 \pm 1.67$\\
         \bottomrule
    \end{tabular}
    \caption{Experiments execution times. Empirical mean and deviation are calculated across 10 runs on ER1 data with fixed method, number of nodes and distribution of the noise terms.}
    \label{tab:exe_time}
\end{table}

\section{Conclusion}
The assumption of Gaussian noise terms in an additive nonlinear model, when violated, can lead causal discovery algorithms to infer graphs with inverted direction of the edges. In this work we prove such limitation, and in response to this problem we introduce NoGAM. Based on the interplay between score matching and causal discovery introduced by \cite{rolland_2022}, our algorithm proposes a novel and consistent method of inference of the topological ordering that doesn't assume any distribution on the noise terms. We prove via systematic experiments that our approach outperforms traditional and state of the art causal discovery algorithms on almost any synthetic dataset generated under arbitrary distribution of the noise terms.

\acks{This work has been supported by  AFOSR,  grant n. FA8655-20-1-7035. FM is supported by \textit{Programma Operativo Nazionale ricerca e innovazione 2014-2020.}}

\bibliography{biblio}

\newpage
\appendix

\section{Identifiability of the multivariate ANM}\label{app:multivariate_anm}
\cite{peters_2014_identifiability} show that Condition \ref{cond:identifiab} for the bivariate model suffices to prove identifiability in the multivariate case. Intuitively, given $d$ structural equations of the form $X_j \coloneqq f_j(\Parents_j) + N_j$ as in model \eqref{eq:scm}, to reproduce a bivariate ANM it is sufficient to fix all arguments of $f_j$ except for one parent $X_i$ and for the noise variable $N_j$. More formally, Definition 27 of \cite{peters_2014_identifiability} define a \textit{restricted additive noise model} as an SCM such that for all nodes  $j \in \{1, \ldots, d\}$, $i \in \Parents_j$, and for all sets $S \subseteq \{1, \ldots, d\}$ where $\Parents_j \setminus \{i\} \subseteq S \subseteq \Nondesc_j \setminus \{i, j\}$, there is a value of $\mathbf{X}_\mathsmaller{S}$ with joint density $p_{\mathsmaller{\mathbf{X}_S}}(\mathbf{x}_\mathsmaller{S}) > 0$ such that the triple $$(f_j(\parents_j \setminus \{i\}, X_i), p_{\mathsmaller{X_i \mid \mathbf{X}_s}}(X_i \mid \mathbf{x}_\mathsmaller{S}), p_\mathsmaller{N_j}(N_j))$$ satisfies Condition \ref{cond:identifiab}. With an abuse of notation we use $\Parents_i$ to denote both the nodes in the causal graph and the random variables associated to them. Also, we denote with $\parents_j$ the observed value of the vector of random variables $\Parents_j$, and in general we rely on upper case notation for random variables and lower case notation for their realizations. \\
Theorem 28 of \cite{peters_2014_identifiability} prove that if a distribution is generated according to a \textit{restricted additive noise model} that satisfies causal minimality (i.e. causal mechanisms $f_j$ non-constant in any of their argument), then the causal graph is identifiable from observational data. We assume model \eqref{eq:scm} to be a \textit{restricted ANM} according to Definition 27 of \cite{peters_2014_identifiability}, ensuring identifiability of the causal graph.

\section{Proof of Lemma \ref{lem:lemma}}\label{ap:lem_proof}
\begin{enumerate}[label=(\roman*)]
\item $X_i$ leaf node $\Rightarrow \Mu\left[\left(h^*(R_i) - s_i(\mathbf{X})\right)^2\right] = 0$: true by Equation \eqref{eq:zero_err}.
\item $\Mu\left[\left(h^*(R_i) - s_i(\mathbf{X})\right)^2\right] = 0 \Rightarrow X_i$ leaf node: 
the zero expectation can be rewritten as $$\mathlarger{\int}_{\R \times \R^{d}} \left(h^*(R_i) - s_i(\mathbf{X})\right)^2 dp(R_i, \mathbf{X}) = 0 \:.$$
Being the integral taken over a positive function, it is immediate that the equality with zero holds if and only if $h^*(R_i) = s_i(\mathbf{X})$ with probability $1$, such that $\mathbb{P}_{\mathbf{X} \mid R_i}(s_i(\mathbf{X}) = h^*(R_i) \mid R_i) = 1$, $\forall \: R_i \in \mathbb{R}$. It follows that the conditional variance satisfies $\Var_{\mathbf{X}}[s_i(\mathbf{X}) \mid R_i] = 0$ for all $R_i \in \mathbb{R}$, and that $s_i(\mathbf{X})$ is almost surely constant given $R_i$ observed. 
We are going to explicit this fact by introducing additional notation. We define $c_{\Rsub} \coloneqq h^*(R_i)$ such that it is clear that $c_{\Rsub}$ is a constant when $R_i$ is fixed.

Now, we are going to prove that $X_i$ must be a leaf for the bivariate case of model \eqref{eq:scm}. Then, we will generalize the arguments to the generic $n$ variables case. 
Let $(X_i, X_j)$ be the nodes of a bivariate graph $\mathcal{G}$. By contradiction, suppose that $X_i$ is not a leaf in the graph. 
By equation \eqref{eq:score_i_generic} the $i$-th entry of the score can be written as
\begin{equation}\label{eq:score_bivariate}
    \begin{split}
        s_i(X_i, X_j) &= \partial_{\mathsmaller{N_i}}g_i(R_i + \mathbf{E}[N_i | X_j]) - \partial_{\mathsmaller{X_i}}f_j(X_i)\partial_{\mathsmaller{N_j}}g_j(X_j - f_j(X_i)) = \\
        &=: g'_i(R_i + \mathbf{E}[N_i | X_j]) - f'_j(X_i)g'_j(X_j - f_j(X_i)) \,.
    \end{split}
\end{equation}

Conditional on $R_i$ the score entry is almost surely constant, i.e. $s_i(X_i, X_j) \mid R_i = c_\Rsub$ with probability $1$.

By identifiability assumption on the generating ANM, $f'_j(x_i)g''_j(x^*_j - f_j(x_i)) \neq 0$ for all but finite $(x_i, x_j) \in \R^2$ (Equation \eqref{eq:f'g''}). Let us define $\mathcal{X}$ as the set uncountable pairs $(x_i, x_j)$ for which such condition is verified, meaning that
\begin{equation*}
    \mathcal{X} \coloneqq \{(x_i, x_j) \in \R^2 \mid f'_j(x_i)g''_j(x_j - f_j(x_i)) \neq 0\} \subset \R^2 \,.
\end{equation*}

Our next goal is to show that, in contradiction with the fact that $\Var_{\mathbf{X}}[s_i(\mathbf{X}) \mid R_i] = 0$ (immediate consequence of the hypothesis of vanishing expectation), there is some value of $R_i$ such that $s_i(x_i, x_j) \neq s_i(x^*_i, x^*_j) \mid R_i$ for distinct pairs $(x_i, x_j)$, $(x^*_i, x^*_j)$ in the support of $\mathbf{X} \mid R_i$. By definition in Equation \eqref{eq:residual} we have $R_i = N_i - \mathbf{E}[N_i \mid X_j]$, and by identifiability assumption of strictly positive density of the noise terms we know that $p_{\mathsmaller{N_i}}(n_i) > 0$ for each supported $n_i$. Then, it is clear that the support of $\mathbf{X}$ and of its transformation $s_i(\mathbf{X})$ is not restricted by the observation of $R_i$: in fact, for each observation $R_i = r_i$, for any value of $X_j=x_j$, $\exists \, n_i \textnormal{ s.t. } p_{N_i}(n_i) > 0$ that allows $r_i = n_i - \mathbf{E}\left[N_i \mid x_j \right]$. 

Therefore we know that there exists an uncountable set of points $(x_i, x_j) \in \mathcal{X}$ that satisfies $s_i(x_i, x_j) = c_{R_i}$ and $f_j'(x_i)g''_j(x_j - f_j(x_i)) \neq 0$, conditional on $R_i$. We also know by Equation \eqref{eq:non_constant_score} that $f_j'(x_i)g''_j(x_j - f_j(x_i)) \neq 0$ implies that $g'_i(x_i) - f_j'(x_i)g'_j(x_j - f_j(x_i))$ has non zero partial derivative on $x_i$ for all pairs $(x_i, x_j) \in \mathcal{X}$, i.e. is never constant on $x_i$. Given that $g'_i(x_i) - f'_j(x_i)g'_j(x_j - f_j(x_i))$ is exactly the analytical expression of the $i$-th score entry, then we have that $s_i(x_i, x_j) \neq s_i(x^*_i, x^*_j) \mid R_i$ for any$(x_i, x_j), (x^*_i, x^*_j) \in \mathcal{X}$ and $x_i \neq x^*_i$, $x_j \neq x^*_j$. Thus, conditional on $R_i$, we have that $\Var_{\mathbf{X}}[s_i(\mathbf{X}) \mid R_i] > 0$, which contradicts the assumption.

Then $X_i$ must be a leaf, which proves the claim of the Lemma for the bivariate case.

\vspace{.5em}
Now, we consider a multivariate  restricted ANM as in \eqref{eq:scm}. Again, by contradiction we assume $X_i$ to be a non-leaf node. Being the model identifiable, we know that for each node $q \in \{1, \ldots, d\}$, $u \in \Parents_q$, and for all sets $S \subseteq \{1, \ldots, d\}$ where $\Parents_q \setminus \{i\} \subseteq S \subseteq \Nondesc_q \setminus \{u, q\}$, there is a value of $\mathbf{x}_\mathsmaller{S}$ with joint density $p_{\mathsmaller{\mathbf{x}_S}}(\mathbf{x}_\mathsmaller{S}) > 0$ such that the triple $$(f_q(\parents_q \setminus \{u\}, X_u), p_{\mathsmaller{X_u \mid \mathbf{X}_s}}(X_u \mid \mathbf{X}_S), p_\mathsmaller{N_q}(N_q))$$ satisfies Condition \ref{cond:identifiab}. 
Let $i_c$ be a children of node $i$ with $i_c \not\in \Parents_k$ for each $k \in \Child_i$. Such node $i_c$ always exists due to the acyclicity constraint on the causal graph. Let $S=\Nondesc_{i_c} \setminus \{i, i_c\}$: conditioning on the set of random variables $\mathbf{X}_{S} = \mathbf{x}_{\mathsmaller{S}}$ , the score entry for node $i$ is 
\begin{equation}\label{eq:score_multivar_proof}
    \begin{split}
            s_i(\mathbf{X}) &= \hspace{.1em}g'_i(R_i + \mathbf{E}[N_i | \child_{i \mathsmaller{\setminus \{X_{i_c}\}}}, X_{i_c}]) + \\
            &- \partial_{\mathsmaller{X_{i}}}f_{i_c}(\parents_{i \mathsmaller{\setminus \{X_{i}\}}}, X_{i})g'_{i_c}(X_{i_c} - f_{i_c}(\parents_{i_c \mathsmaller{\setminus \{X_{i}\}}}, X_{i})) + \\
            &- \sum_{k \in \Child_{i \setminus \{i_c\}}} \partial_{\mathsmaller{X_i}}f_k(\parents_{k \mathsmaller{\setminus \{X_{i}\}}}, X_{i})\partial_{\mathsmaller{N_k}} g_k(n_k = x_k - f_k(\parents_{k \mathsmaller{\setminus \{X_{i}\}}}, X_{i}))
    \end{split}
\end{equation}
If we fix $X_i = x^*_i$ such that for all the uncountable pairs $(x^*_i, x_{i_c})$ such that $g'_i(N_i) - \partial_{\mathsmaller{X_{i}}}f_{i_c}(\parents_{i \mathsmaller{\setminus \{X_{i}\}}}, x^*_i)g'_{i_c}(X_{i_c} - f_{i_c}(\parents_{i \mathsmaller{\setminus \{X_{i}\}}}, x^*_i))$ is not constant, then we can observe that: the first two terms of Equation \eqref{eq:score_multivar_proof} are exactly analogous to Equation \eqref{eq:score_bivariate} in the bivariate case, for which non vanishing variance under observation of $R_i$ is proven; the summation on the remaining children instead does not contribute to the variance under conditioning on $\mathbf{x}_S$ and $x^*_i$. Thus, we know that $\Var_{\mathbf{X}}[s_i(\mathbf{X}) \mid R_i] > 0$, which contradicts the assumption. We conclude that $X_i$ must be a leaf.

\end{enumerate}

\section{Pruning of the DAG with CAM}\label{app:cam}
Once the DAG constraint is enforced by a topological order, our NoGAM algorithm removes superfluous edges from the graph using \textit{CAM-pruning} procedure 
(\cite{B_hlmann_2014}). In fact, a smaller DAG typically yields to statistically more efficient estimations of interventional distributions with respect to the fully connected graph compatible with the topological order. The idea is that, under assumption of additive structure of $f_i$ nonlinear mechanisms in \eqref{eq:scm}, one can perform regression on potential parent nodes and use additive hypothesis testing (\cite{marra_feature_selection}) to decide about the presence of an edge. For more details, please refer to the original paper of \cite{B_hlmann_2014}.

\section{Comparison with SCORE}\label{app:score}
In this Section we provide a summary of the key differences of our work with respect to \cite{rolland_2022}, and a brief introduction to the Stein gradient estimator, the main ingredient common to the implementation of SCORE and NoGAM algorithms. \cite{rolland_2022} proposes a method for identification of leaf nodes by inspection of the diagonal elements of the Jacobian of the score function: this is done identifying nonlinearities in the diagonal entries by estimation of their variance, which is possible only under assumption of Gaussian noise terms. Additionally, in order to guarantee nonlinearities in Jacobian of the score, it is required that the causal mechanisms $f_i$ are nonlinear in each of their arguments: this is a strong assumption which violation leads SCORE to infer the wrong topological ordering. To better clarify this point, we explicitly consider the analytical form of a diagonal entry of the Jacobian of the score function:
\begin{equation*}
    \partial_{X_i}s_i(\mathbf{X}) = \partial^2_{\mathsmaller{X_i}} \log p_i(X_i \mid \Parents_i) + \sum_{k \in \Child_i} \partial_{\mathsmaller{X_i}}(\partial_{\mathsmaller{X_i}}f_k(\Parents_k)\partial_{\mathsmaller{X_k}} \log p_k(X_k \mid \Parents_k)) \; .
\end{equation*}
By writing $p_k(X_k \mid \Parents_k)$ normal distribution explicitly, it is easy to see how terms in the summation over the children are vanishing if $f_k$ is linear in $X_i$ (due to the second order partial derivative of $f_k(\Parents_k)$).
Our paper instead develops the theory to identify leaf nodes in a causal graph with less restrictive assumptions, both on the noise and on the functional mechanisms, based on minimization of the generalization error in the prediction of the score entries from the residuals of Equation \eqref{eq:residual}.\\[.4em]
From a practical viewpoint, both SCORE and NoGAM methods rely on efficient approximation of $\nabla_{\mathbf{X}}\log p(\mathbf{X})$ using the score matching based Stein gradient estimator (\cite{stein_gradient}), which we briefly outline below.
\paragraph{Stein gradient estimator} The estimator is based on the Stein identity, that was first developed for Gaussian random variables (\cite{Stein1972ABF}) and then extended to the general case (\cite{gorham17a_stein, liu106_stein}). For any test function $\mathbf{h} : \R^d \rightarrow \R^{d'}$ such that $\lim_{\mathbf{x} \rightarrow \mathbf{\infty}}\mathbf{h}(\mathbf{x}))p(\mathbf{x}) = 0$, the following identity holds:
$$\mathbf{E}_p\left[\mathbf{h}(\mathbf{x})\nabla_{\mathbf{x}}\log p(\mathbf{x}) + \nabla_{\mathbf{x}}\mathbf{h}(\mathbf{x}) \right] = 0.$$
We note that the quantity of interest  $\nabla_{\mathbf{x}}\log p(\mathbf{x})$ appears in the expression. Being the integral of the expectation intractable, \cite{stein_gradient} propose to exploit Monte Carlo approximation of the expectation, and then show that it is possible to derive an estimator of the score consistent in the large sample limit. For additional details, please refer to the original manuscript.

\section{Discussion on Example \ref{inversion_example}}\label{app:example}
Example \ref{inversion_example} experimentally illustrates the shortcomings of inference under assumption of Gaussian noise terms, when this is violated in the ground truth generative model. In what follows we are going to provide a more detailed overview of the experimental design of the example and of the method of inference of the causal effect direction. For both the linear and nonlinear settings we generate $2000$ samples of the ground truth noise terms $X_1$ and $N_2$, uniformly distributed with support in the $[0, 5]$ and $[1, 4]$ intervals, respectively (see Figure \ref{fig:x1_eps2}). Theorem 1 of \cite{Zhang09_additive} proves that when fitting additive model \eqref{eq:scm} with the causal structure represented by a DAG, then the total entropy of the disturbances, i.e. $\sum_i^d H(N_i)$, is minimized at the minimum of $H(X_1, \ldots, X_d)$ (which corresponds to the minimum in the negative log-likelihood). Thus, in the bivariate case of Example \ref{inversion_example}, we can choose as the correct causal direction the one achieving the minimum in the total entropy of the estimated noise terms.

\section{Causal discovery with independence of estimated residuals}\label{app:comparison_mooij_peters}
\cite{mooij2009} and \cite{peters_2014_identifiability} propose a causal discovery methodology comparable to our Algorithm \ref{alg:top_order}, as they operate on identifiable ANM without posing restrictions on the distribution of the noise terms. In practice, they estimate the causal  structure in an iterative way, performing regression similarly to our Equation \eqref{eq:res_problem} and testing for independence of the inferred residuals $R_i$ and variables $\Xminus{i}$. The main bottleneck of this approach is the use of independence testing, which is hard to perform (\cite{shah2018_ci_test}) as well as to scale to high dimensional graphs and  large size datasets. Additionally, it does not ensure consistency of the inferred graph in the population case, unless an oracle independence test is assumed. In Table \ref{tab:resit_experiments} we compare empirical performance of NoGAM and RESIT method of \cite{peters_2014_identifiability}. We reproduce the experimental setting of \cite{peters_2014_identifiability} (Section 5.1.2), sampling the nonlinear mechanisms from a Gaussian process with unitary bandwidth and independent noise terms under normal distribution and variance uniformly chosen. 
Experiments are run with number of samples  $n \in \{100, 500\}$ and number of variables $d \in \{4, 15\}$, in a sparse setting. We can see that NoGAM outperforms RESIT under any of the experimental configurations.

\begin{table}[htb!]
\footnotesize
\caption{Experimental performance of NoGAM compared to RESIT. RESIT results are taken from \cite{peters_2014_identifiability}, Table 3 and Table 4.}
\label{tab:resit_experiments}
\centering
\begin{tabular}{llcc}
\toprule
 & Method & SHD (n=100) & SHD (n=500)\\
\midrule
\multirow{2}{*}{d=4}   & \textbf{NoGAM} & $\mathbf{1.6 \pm 1.1}$ & $\mathbf{0.4 \pm 0.4}$\\
& RESIT & $1.7 \pm 1.3$ & $0.8 \pm 0.9$\\
\midrule
\multirow{2}{*}{d=15}  & \textbf{NoGAM} & $\mathbf{11.7 \pm 2.4}$ & $\mathbf{7.6 \pm 4.1}$\\
& RESIT & $15.4 \pm 5.7$ & $10.1 \pm 5.7$\\
\bottomrule
\end{tabular}
\end{table}


%
\section{Experiments on ER graphs}\label{app:er}
\begin{figure}
     \centering
     \begin{subfigure}[b]{0.32\textwidth}
         \centering
         \includegraphics[width=\textwidth]{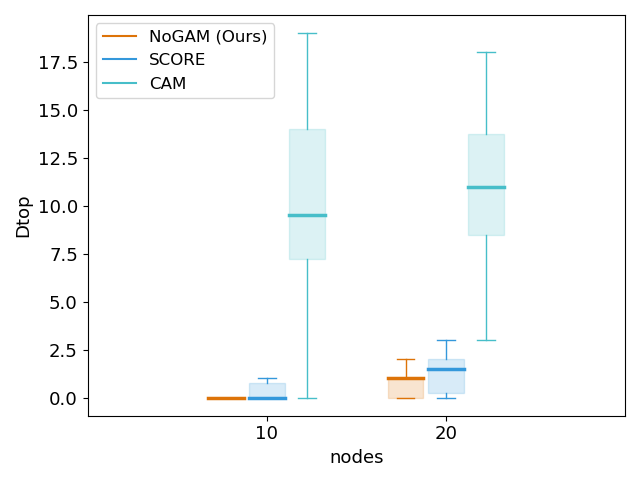}
         \caption{Beta distribution}
     \end{subfigure}%
     \hfill
     \begin{subfigure}[b]{0.32\textwidth}
         \centering
         \includegraphics[width=\textwidth]{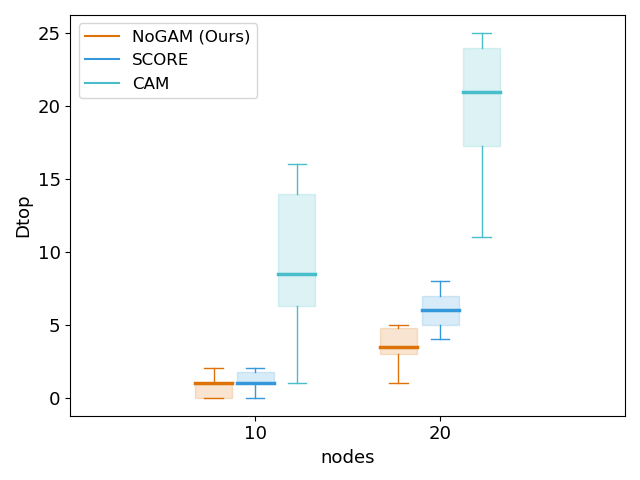}
         \caption{Exponential distribution}
     \end{subfigure}%
     \hfill
     \begin{subfigure}[b]{0.32\textwidth}
         \centering
         \includegraphics[width=\textwidth]{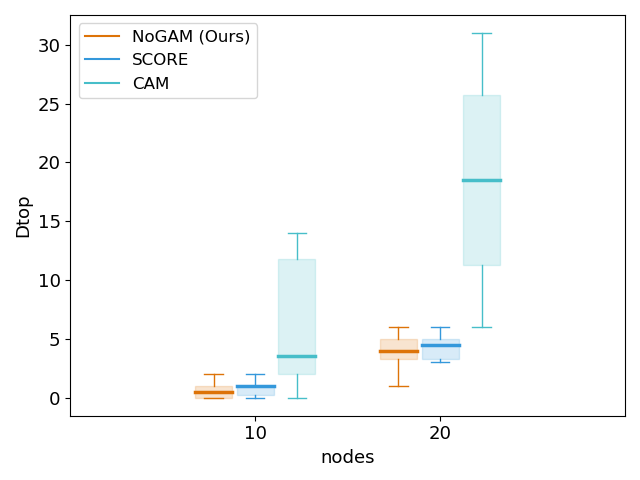}
         \caption{Gamma distribution}
     \end{subfigure}%
     
     \medskip
          \begin{subfigure}[b]{0.32\textwidth}
         \centering
         \includegraphics[width=\textwidth]{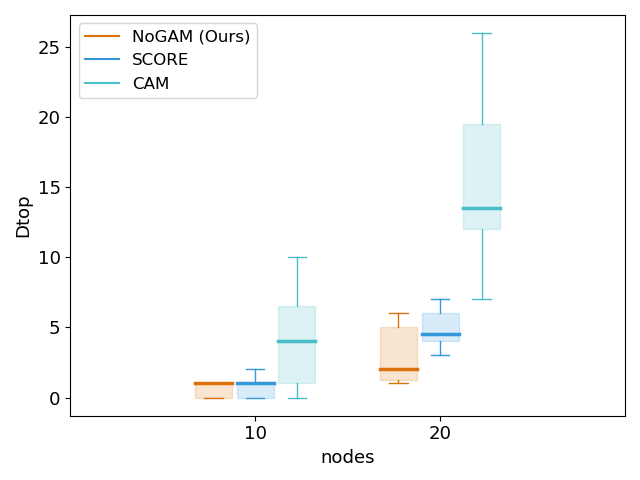}
         \caption{Normal distribution}
     \end{subfigure}%
     \hfill
     \begin{subfigure}[b]{0.32\textwidth}
         \centering
         \includegraphics[width=\textwidth]{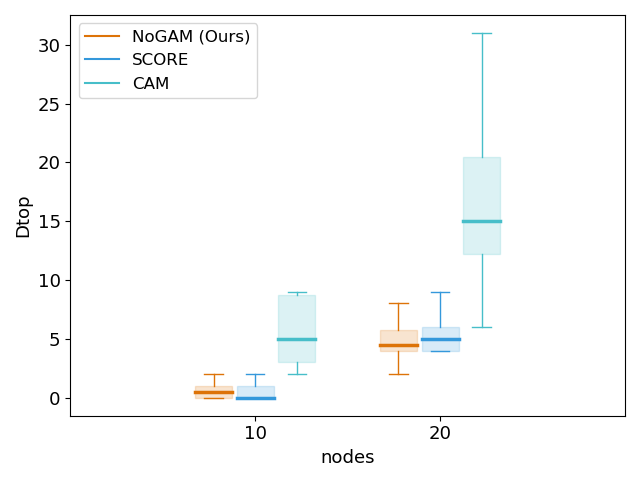}
         \caption{Gumbel distribution}
     \end{subfigure}%
     \hfill
     \begin{subfigure}[b]{0.32\textwidth}
         \centering
         \includegraphics[width=\textwidth]{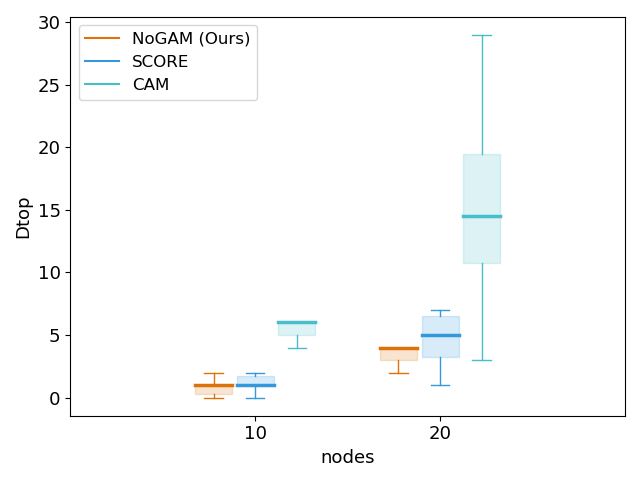}
         \caption{Laplace distribution}
     \end{subfigure}
        \caption{Boxplots over 10 runs for topological ordering divergence $D_{top}$ over dense graphs (ER4). GES algorithm doesn't appear since it does not require an explicit estimate of the topological ordering. From the illustrations we can see how NoGAM, in general, outperforms CAM and SCORE.} 
        \label{fig:dtop_boxplots_er4}
\end{figure}

\begin{figure}
     \centering
     \begin{subfigure}[b]{0.32\textwidth}
         \centering
         \includegraphics[width=\textwidth]{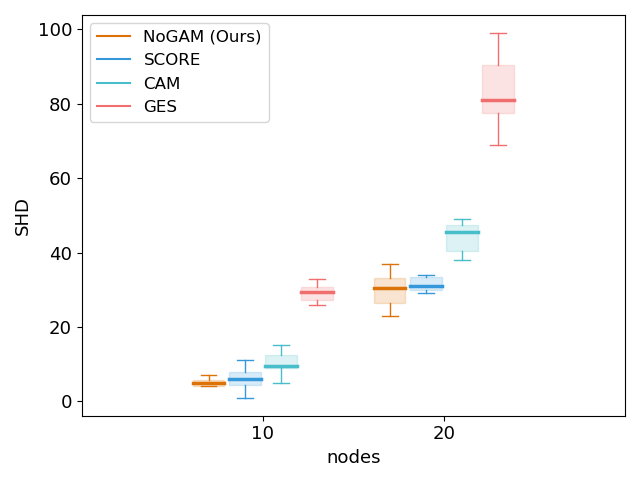}
         \caption{Beta distribution}
     \end{subfigure}%
     \hfill
     \begin{subfigure}[b]{0.32\textwidth}
         \centering
         \includegraphics[width=\textwidth]{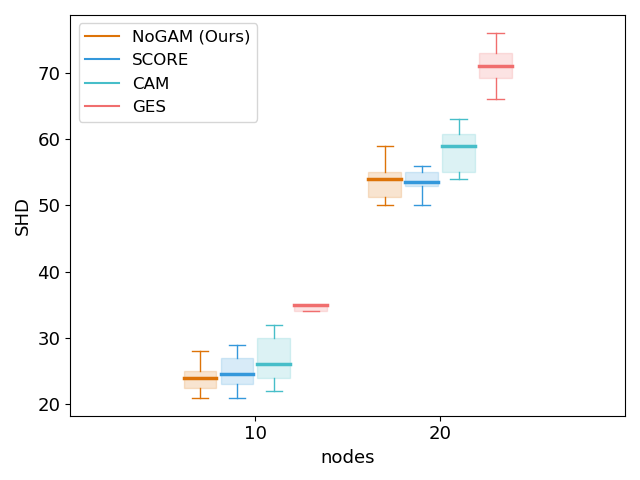}
         \caption{Exponential distribution}
     \end{subfigure}%
     \hfill
     \begin{subfigure}[b]{0.32\textwidth}
         \centering
         \includegraphics[width=\textwidth]{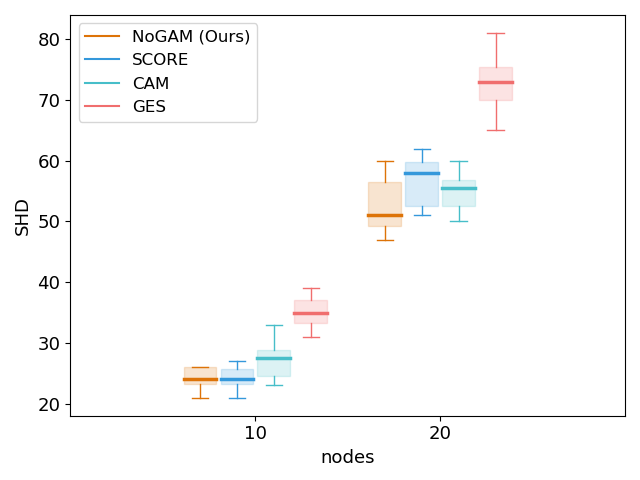}
         \caption{Gamma distribution}
     \end{subfigure}%
     
     \medskip
          \begin{subfigure}[b]{0.32\textwidth}
         \centering
         \includegraphics[width=\textwidth]{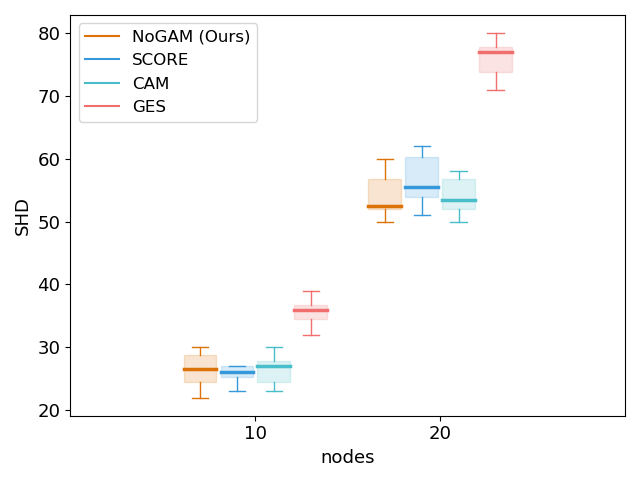}
         \caption{Normal distribution}
     \end{subfigure}%
     \hfill
     \begin{subfigure}[b]{0.32\textwidth}
         \centering
         \includegraphics[width=\textwidth]{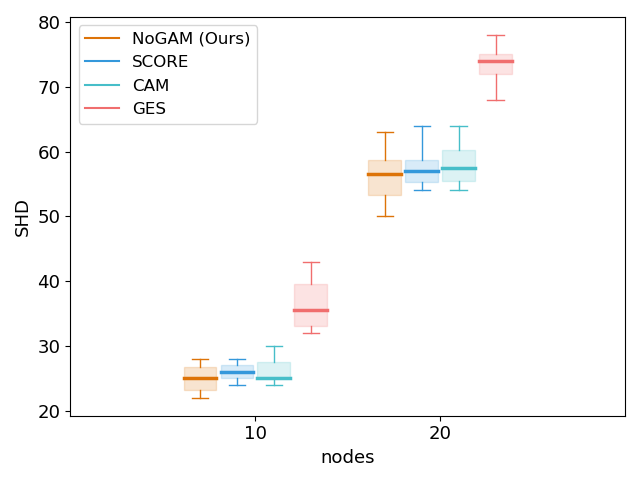}
         \caption{Gumbel distribution}
     \end{subfigure}%
     \hfill
     \begin{subfigure}[b]{0.32\textwidth}
         \centering
         \includegraphics[width=\textwidth]{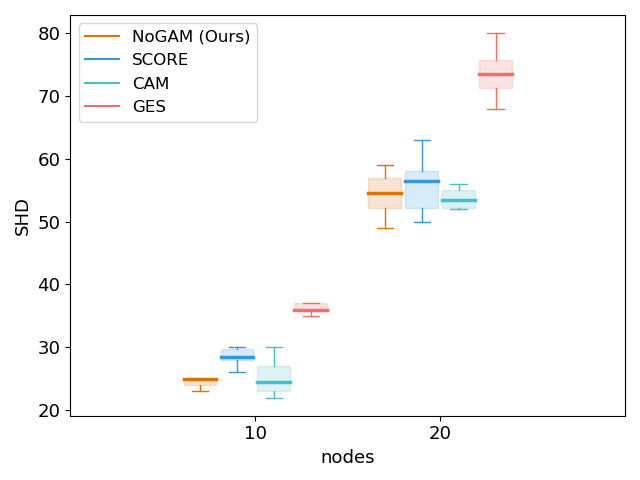}
         \caption{Laplace distribution}
     \end{subfigure}
        \caption{Boxplots over 10 runs showing SHD performance over dense graphs (ER4).}
        \label{fig:shd_er4}
\end{figure}

From Table \ref{tab:beta} to \ref{tab:laplace} we provide the complete overview of the experiments described in section \ref{sec:experiments}. Each metric is averaged over $10$ runs, for which we record the empirical mean and deviation. Figure \ref{fig:dtop_boxplots_er4} shows how in the dense setting (ER4) NoGAM method of inference of the topological ordering outperforms CAM and SCORE, similarly to what observed in the sparse setting (ER1) illustrated in Figure \ref{fig:dtop_boxplots_er1}. In Figure \ref{fig:shd_er4} we can see that SCORE and NoGAM have overall comparable SHD.

\begin{table}[htb!]
\footnotesize
\caption{Beta noise ER graphs}
\label{tab:beta}
\centering
\begin{tabular}{clcccccc}
\toprule
\multicolumn{2}{c}{} & \multicolumn{3}{l}{ER1 (sparse)} & \multicolumn{3}{l}{ER4 (dense)}\\
\midrule
 & Method & SHD & SID & $D_{top}$ & SHD & SID & $D_{top}$\\
\midrule
\multirow{4}{*}{d=10}   & NoGAM & $1.1 \pm 1.5$ & $4.2 \pm 5.4$ & $0.6 \pm 0.8$  &  $5.1 \pm 1.3$ & $11.4 \pm 7.1$ & $0.1 \pm 0.3$ \\
& SCORE & $1.6 \pm 1.0$ & $4.9 \pm 3.5$ & $0.6 \pm 0.7$  &  $6.1 \pm 2.0$ & $9.6 \pm 5.4$ & $0.4 \pm 0.7$ \\
& CAM & $3.4 \pm 1.9$ & $13.8 \pm 8.6$ & $2.2 \pm 1.4$  &  $14.4 \pm 5.1$ & $51.0 \pm 14.3$ & $7.7 \pm 5.1$ \\
& GES & $6.5 \pm 2.0$ & $22.0 \pm 7.6$ & $-$  &  $27.0 \pm 5.2$ & $72.0 \pm 8.7$ & $-$ \\
\midrule
\multirow{4}{*}{d=20}  & NoGAM & $2.0 \pm 1.6$ & $8.0 \pm 9.0$ & $0.5 \pm 0.5$  &  $30.0 \pm 4.2$ & $121.5 \pm 22.6$ & $0.9 \pm 0.9$ \\
& SCORE &  $2.0 \pm 1.5$ & $8.0 \pm 5.2$ & $1.0 \pm 0.3$  &  $33.1 \pm 6.8$ & $111.9 \pm 29.1$ & $1.4 \pm 1.0$ \\
& CAM & $9.5 \pm 4.8$ & $45.3 \pm 16.0$ & $5.7 \pm 2.3$  &  $45.1 \pm 7.2$ & $220.8 \pm 29.1$ & $12.2 \pm 7.0$ \\
& GES & $12.9 \pm 4.7$ & $52.0 \pm 14.4$ & $-$  &  $78.5 \pm 9.2$ & $290.6 \pm 25.9$ & $-$ \\
\bottomrule
\end{tabular}
\end{table}

\begin{table}[htb!]
\footnotesize
\caption{Exponential noise ER graphs}
\label{tab:exponential}
\centering
\begin{tabular}{clcccccc}
\toprule
\multicolumn{2}{c}{} & \multicolumn{3}{l}{ER1 (sparse)} & \multicolumn{3}{l}{ER4 (dense)}\\
\midrule
 & Method & SHD & SID & $D_{top}$ & SHD & SID & $D_{top}$\\
\midrule
\multirow{4}{*}{d=10}   & NoGAM & $1.2 \pm 1.4$ & $4.3 \pm 5.2$ & $0.3 \pm 0.6$  &  $24.1 \pm 2.0$ & $43.3 \pm 5.2$ & $0.8 \pm 0.8$ \\
& SCORE & $2.3 \pm 1.2$ & $4.1 \pm 4.2$ & $0.8 \pm 0.9$  &  $25.1 \pm 3.0$ & $40.9 \pm 6.6$ & $1.1 \pm 0.6$ \\
& CAM & $3.4 \pm 1.4$ & $14.8 \pm 5.8$ & $2.4 \pm 1.0$  &  $24.6 \pm 2.0$ & $54.1 \pm 11.9$ & $4.4 \pm 2.1$ \\
& GES & $9.0 \pm 3.0$ & $33.1 \pm 12.2$ & $-$  &  $33.9 \pm 3.7$ & $80.6 \pm 5.8$ & $-$ \\
\midrule
\multirow{4}{*}{d=20}  & NoGAM & $2.5 \pm 1.4$ & $11.3 \pm 9.1$ & $0.7 \pm 0.6$  &  $54.2 \pm 4.5$ & $218.1 \pm 28.7$ & $3.6 \pm 1.6$ \\
& SCORE &  $4.2 \pm 2.8$ & $19.7 \pm 13.6$ & $1.4 \pm 1.4$  &  $54.1 \pm 4.0$ & $215.1 \pm 40.4$ & $6.1 \pm 1.3$ \\
& CAM & $8.2 \pm 1.8$ & $46.4 \pm 24.4$ & $5.5 \pm 2.1$  &  $54.8 \pm 4.3$ & $258.4 \pm 57.4$ & $18.8 \pm 7.5$ \\
& GES & $19.8 \pm 3.5$ & $87.8 \pm 24.4$ & $-$  &  $72.9 \pm 5.4$ & $336.2 \pm 24.0$ & $-$ \\
\bottomrule
\end{tabular}
\end{table}

\begin{table}[htb!]
\footnotesize
\caption{Gamma noise ER graphs}
\label{tab:gamma}
\centering
\begin{tabular}{clcccccc}
\toprule
\multicolumn{2}{c}{} & \multicolumn{3}{l}{ER1 (sparse)} & \multicolumn{3}{l}{ER4 (dense)}\\
\midrule
 & Method & SHD & SID & $D_{top}$ & SHD & SID & $D_{top}$\\
\midrule
\multirow{4}{*}{d=10}   & NoGAM & $1.2 \pm 1.2$ & $3.9 \pm 5.4$ & $0.3 \pm 0.5$  &  $24.2 \pm 1.8$ & $41.4 \pm 5.3$ & $0.7 \pm 0.8$ \\
& SCORE & $1.6 \pm 1.5$ & $4.8 \pm 6.3$ & $1.4 \pm 0.9$  &  $24.2 \pm 0.9$ & $41.6 \pm 2.5$ & $1.0 \pm 1.1$ \\
& CAM & $3.3 \pm 1.6$ & $17.3 \pm 8.2$ & $2.6 \pm 1.7$  &  $26.5 \pm 3.8$ & $59.3 \pm 15.1$ & $8.0 \pm 4.5$ \\
& GES & $9.8 \pm 2.2$ & $34.6 \pm 13.2$ & $-$  &  $35.9 \pm 3.6$ & $80.8 \pm 6.5$ & $-$ \\
\midrule
\multirow{4}{*}{d=20}  & NoGAM & $3.9 \pm 1.7$ & $13.5 \pm 10.1$ & $0.7 \pm 0.8$  &  $52.7 \pm 2.8$ & $203.1 \pm 24.9$ & $3.7 \pm 1.6$ \\
& SCORE &  $4.0 \pm 1.8$ & $14.8 \pm 8.2$ & $2.5 \pm 1.3$  &  $56.5 \pm 2.3$ & $214.7 \pm 21.7$ & $4.7 \pm 1.7$ \\
& CAM & $8.3 \pm 3.0$ & $48.9 \pm 23.3$ & $5.1 \pm 2.5$  &  $54.0 \pm 3.5$ & $264.5 \pm 17.6$ & $16.2 \pm 4.6$ \\
& GES & $18.8 \pm 2.5$ & $89.0 \pm 19.4$ & $-$  &  $73.9 \pm 3.2$ & $332.8 \pm 19.2$ & $-$ \\
\bottomrule
\end{tabular}
\end{table}

\begin{table}[htb!]
\footnotesize
\caption{Gauss noise ER graphs}
\label{tab:gaussian}
\centering
\begin{tabular}{clcccccc}
\toprule
\multicolumn{2}{c}{} & \multicolumn{3}{l}{ER1 (sparse)} & \multicolumn{3}{l}{ER4 (dense)}\\
\midrule
 & Method & SHD & SID & $D_{top}$ & SHD & SID & $D_{top}$\\
\midrule
\multirow{4}{*}{d=10}   & NoGAM &  $0.4 \pm 0.5$ & $0.0 \pm 0.0$ & $0.0 \pm 0.0$  &  $26.4 \pm 2.5$ & $43.9 \pm 4.8$ & $0.8 \pm 0.9$ \\
& SCORE & $0.9 \pm 0.7$ & $3.2 \pm 2.4$ & $0.1 \pm 0.3$  &  $26.1 \pm 3.0$ & $43.7 \pm 8.7$ & $0.7 \pm 0.6$ \\
& CAM & $0.6 \pm 0.7$ & $1.3 \pm 1.6$ & $0.0 \pm 0.0$  &  $26.9 \pm 2.2$ & $47.4 \pm 5.2$ & $5.1 \pm 3.4$ \\
& GES & $8.3 \pm 1.8$ & $31.9 \pm 9.5$ & $-$  &  $36.2 \pm 2.4$ & $85.3 \pm 3.8$ & $-$ \\
\midrule
\multirow{4}{*}{d=20}  & NoGAM & $2.0 \pm 1.0$ & $10.1 \pm 5.7$ & $0.3 \pm 0.5$  &  $54.0 \pm 3.0$ & $195.0 \pm 17.0$ & $3.1 \pm 2.0$ \\
& SCORE & $2.3 \pm 1.3$ & $11.6 \pm 6.9$ & $0.4 \pm 0.7$  &  $56.5 \pm 2.9$ & $197.8 \pm 22.3$ & $5.0 \pm 1.3$ \\
& CAM & $2.1 \pm 1.8$ & $9.8 \pm 9.9$ & $0.8 \pm 1.0$  &  $57.1 \pm 1.9$ & $204.1 \pm 34.2$ & $21.4 \pm 8.4$ \\
& GES & $17.1 \pm 3.7$ & $70.0 \pm 22.1$ & $-$  &  $71.9 \pm 3.8$ & $340.1 \pm 20.9$ & $-$ \\
\bottomrule
\end{tabular}
\end{table}

\begin{table}[htb!]
\footnotesize
\caption{Gumbel noise ER graphs}
\label{tab:gumbel}
\centering
\begin{tabular}{clcccccc}
\toprule
\multicolumn{2}{c}{} & \multicolumn{3}{l}{ER1 (sparse)} & \multicolumn{3}{l}{ER4 (dense)}\\
\midrule
 & Method & SHD & SID & $D_{top}$ & SHD & SID & $D_{top}$\\
\midrule
\multirow{4}{*}{d=10}   & NoGAM &  $1.3 \pm 1.4$ & $3.3 \pm 4.2$ & $0.2 \pm 0.4$  &  $24.9 \pm 2.7$ & $41.4 \pm 4.8$ & $0.6 \pm 0.7$ \\
& SCORE & $1.6 \pm 1.4$ & $6.3 \pm 5.2$ & $0.5 \pm 0.5$  &  $26.1 \pm 2.1$ & $42.6 \pm 5.7$ & $0.5 \pm 0.7$ \\
& CAM & $1.4 \pm 1.3$ & $7.9 \pm 8.8$ & $0.8 \pm 1.1$  &  $26.5 \pm 1.9$ & $55.9 \pm 12.4$ & $7.1 \pm 4.3$ \\
& GES & $9.7 \pm 1.2$ & $41.0 \pm 11.1$ & $-$  &  $38.2 \pm 1.3$ & $86.8 \pm 2.5$ & $-$ \\
\midrule
\multirow{4}{*}{d=20}  & NoGAM & $2.7 \pm 2.4$ & $11.3 \pm 6.8$ & $0.6 \pm 0.7$  &  $56.1 \pm 3.9$ & $207.9 \pm 20.4$ & $4.7 \pm 2.4$ \\
& SCORE & $3.5 \pm 1.6$ & $15.9 \pm 17.4$ & $1.3 \pm 1.3$  &  $56.9 \pm 2.2$ & $205.6 \pm 25.0$ & $5.4 \pm 1.6$ \\
& CAM & $4.3 \pm 1.4$ & $29.7 \pm 12.3$ & $2.3 \pm 1.4$  &  $57.0 \pm 4.3$ & $225.5 \pm 24.7$ & $17.6 \pm 6.3$ \\
& GES & $21.1 \pm 1.9$ & $95.4 \pm 28.5$ & $-$  &  $78.2 \pm 2.4$ & $348.5 \pm 15.9$ & $-$ \\
\bottomrule
\end{tabular}
\end{table}

\begin{table}[htb!]
\footnotesize
\caption{Laplace noise ER graphs}
\label{tab:laplace}
\centering
\begin{tabular}{clcccccc}
\toprule
\multicolumn{2}{c}{} & \multicolumn{3}{l}{ER1 (sparse)} & \multicolumn{3}{l}{ER4 (dense)}\\
\midrule
 & Method & SHD & SID & $D_{top}$ & SHD & SID & $D_{top}$\\
\midrule
\multirow{4}{*}{d=10}   & NoGAM & $0.6 \pm 0.7$ & $2.2 \pm 2.9$ & $0.0 \pm 0.3$  &  $24.5 \pm 2.0$ & $42.9 \pm 6.5$ & $0.8 \pm 0.9$ \\
& SCORE & $1.4 \pm 1.0$ & $3.0 \pm 3.0$ & $0.3 \pm 0.4$  &  $27.9 \pm 2.3$ & $46.5 \pm 6.1$ & $1.1 \pm 0.5$ \\
& CAM &  $1.1 \pm 1.0$ & $4.6 \pm 4.9$ & $0.4 \pm 0.5$  &  $24.7 \pm 1.7$ & $43.8 \pm 4.3$ & $6.0 \pm 3.1$ \\
& GES & $10.0 \pm 1.7$ & $37.5 \pm 10.1$ & $-$  &  $35.6 \pm 2.4$ & $85.0 \pm 2.3$ & $-$ \\
\midrule
\multirow{4}{*}{d=20}  & NoGAM & $2.0 \pm 2.2$ & $9.1 \pm 6.4$ & $0.4 \pm 0.7$  &  $54.5 \pm 3.4$ & $204.2 \pm 20.1$ & $3.9 \pm 1.5$ \\
& SCORE & $3.1 \pm 1.3$ & $16.1 \pm 6.6$ & $1.3 \pm 0.6$  &  $56.1 \pm 1.9$ & $209.6 \pm 21.4$ & $4.6 \pm 2.0$ \\
& CAM & $3.4 \pm 1.2$ & $11.1 \pm 9.1$ & $0.5 \pm 0.7$  &  $54.8 \pm 2.6$ & $200.4 \pm 16.1$ & $18.5 \pm 6.5$ \\
& GES & $19.7 \pm 2.4$ & $90.4 \pm 32.0$ & $-$  &  $73.4 \pm 3.1$ & $329.4 \pm 13.9$ & $-$ \\
\bottomrule
\end{tabular}
\end{table}


\section{Additional experiments}\label{app:additional_experiments}

\subsection{Experiments on Sachs data}\label{app:sachs}
We test NoGAM on Sachs dataset (\cite{sachs_2005}), a common causal discovery benchmark made of real-world biological data. In the results of  Table \ref{tab:sachs}, we can see how NoGAM retains state of the art performance with respect to the alternative methods.

\begin{table}[]
    \caption{Experimental results on Sachs dataset (11 variables, 17 edges, 853 observables).}
    \label{tab:sachs}
    \centering
    \begin{tabular}{cccc}
        \midrule
         Method & $D_{top}$ & SHD & SID \\
         \midrule
         NoGAM & $8$ & $12$ & $45$ \\
         SCORE & $8$ & $12$ & $45$ \\
         CAM & $7$ & $12$ & $55$ \\
         GES & $-$ & $17$ & $62$\\
         \bottomrule
    \end{tabular}
\end{table}


\subsection{Experiments on Scale Free graphs}\label{app:sf}
From Table \ref{tab:beta_sf} to \ref{tab:laplace_sf} we provide additional experimental results on Scale Free (SF) graphs (\cite{Barabasi99emergenceScaling}). Datasets are generated with $1000$ samples, number of nodes equals $10$ and $20$, average number of edges per node equals $1$ (sparse graphs) and $4$ (dense graphs). 

\begin{table}[htb!]
\footnotesize
\caption{Beta noise SF graphs}
\label{tab:beta_sf}
\centering
\begin{tabular}{clcccccc}
\toprule
\multicolumn{2}{c}{} & \multicolumn{3}{l}{SF1 (sparse)} & \multicolumn{3}{l}{SF4 (dense)}\\
\midrule
 & Method & SHD & SID & $D_{top}$ & SHD & SID & $D_{top}$\\
\midrule
\multirow{4}{*}{d=10}   & NoGAM & $2.6 \pm 2.1$ & $13.7 \pm 15.9$ & $1.7 \pm 1.6$  &  $7.3 \pm 2.3$ & $12.9 \pm 9.1$ & $0.6 \pm 0.7$ \\
& SCORE & $2.3 \pm 1.4$ & $4.5 \pm 4.0$ & $0.7 \pm 0.8$  &  $6.2 \pm 1.4$ & $16.0 \pm 8.3$ & $0.3 \pm 0.5$ \\
& CAM & $4.5 \pm 1.2$ & $29.9 \pm 6.7$ & $3.3 \pm 0.8$  &  $14.3 \pm 4.0$ & $47.7 \pm 18.6$ & $6.4 \pm 4.5$ \\
& GES & $9.8 \pm 2.9$ & $56.3 \pm 17.4$ & $-$  &  $27.6 \pm 5.8$ & $73.7 \pm 5.4$ & $-$ \\
\midrule
\multirow{4}{*}{d=20}  & NoGAM & $2.5 \pm 1.7$ & $24.8 \pm 22.0$ & $0.9 \pm 1.0$  &  $31.1 \pm 6.9$ & $60.3 \pm 20.0$ & $0.8 \pm 0.6$ \\
& SCORE &  $2.1 \pm 1.4$ & $5.8 \pm 3.9$ & $0.1 \pm 0.3$  &  $31.6 \pm 5.8$ & $108.4 \pm 18.7$ & $1.2 \pm 1.2$ \\
& CAM & $6.3 \pm 2.8$ & $84.1 \pm 40.8$ & $4.6 \pm 2.1$  &  $37.7 \pm 7.4$ & $178.9 \pm 44.4$ & $6.8 \pm 4.2$ \\
& GES & $21.3 \pm 6.9$ & $157.8 \pm 48.5$ & $-$  &  $93.3 \pm 13.8$ & $303.2 \pm 13.1$ & $-$ \\
\bottomrule
\end{tabular}
\end{table}

\begin{table}[htb!]
\footnotesize
\caption{Exponential noise SF graphs}
\label{tab:exponential_sf}
\centering
\begin{tabular}{clcccccc}
\toprule
\multicolumn{2}{c}{} & \multicolumn{3}{l}{SF1 (sparse)} & \multicolumn{3}{l}{SF4 (dense)}\\
\midrule
 & Method & SHD & SID & $D_{top}$ & SHD & SID & $D_{top}$\\
\midrule
\multirow{4}{*}{d=10}   & NoGAM & $1.5 \pm 1.1$ & $7.7 \pm 6.7$ & $0.9 \pm 0.8$  &  $8.3 \pm 2.1$ & $31.6 \pm 6.0$ & $0.6 \pm 0.8$ \\
& SCORE & $0.7 \pm 0.8$ & $2.4 \pm 3.6$ & $0.3 \pm 0.6$  &  $8.0 \pm 2.6$ & $39.4 \pm 6.1$ & $0.4 \pm 0.5$ \\
& CAM & $3.6 \pm 2.0$ & $26.6 \pm 12.6$ & $2.8 \pm 1.3$  &  $11.7 \pm 3.3$ & $54.0 \pm 9.1$ & $4.8 \pm 4.0$ \\
& GES & $11.7 \pm 2.6$ & $59.0 \pm 10.7$ & $-$  &  $26.2 \pm 3.0$ & $81.0 \pm 5.9$ & $-$ \\
\midrule
\multirow{4}{*}{d=20}  & NoGAM & $5.3 \pm 2.5$ & $61.7 \pm 28.5$ & $3.6 \pm 1.7$  &  $24.5 \pm 3.9$ & $233.3 \pm 24.4$ & $4.8 \pm 2.4$ \\
& SCORE &  $3.4 \pm 1.7$ & $14.8 \pm 7.0$ & $1.3 \pm 0.9$  &  $24.7 \pm 3.7$ & $209.4 \pm 29.3$ & $4.2 \pm 1.2$ \\
& CAM & $7.7 \pm 2.6$ & $114.0 \pm 37.8$ & $5.8 \pm 1.8$  &  $29.2 \pm 3.2$ & $271.5 \pm 38.4$ & $10.1 \pm 5.2$ \\
& GES & $26.8 \pm 3.5$ & $203.5 \pm 48.5$ & $-$  &  $59.0 \pm 4.1$ & $350.1 \pm 10.6$ & $-$ \\
\bottomrule
\end{tabular}
\end{table}

\begin{table}[htb!]
\footnotesize
\caption{Gamma noise SF graphs}
\label{tab:gamma_sf}
\centering
\begin{tabular}{clcccccc}
\toprule
\multicolumn{2}{c}{} & \multicolumn{3}{l}{SF1 (sparse)} & \multicolumn{3}{l}{SF4 (dense)}\\
\midrule
 & Method & SHD & SID & $D_{top}$ & SHD & SID & $D_{top}$\\
\midrule
\multirow{4}{*}{d=10}   & NoGAM & $0.7 \pm 1.2$ & $6.0 \pm 12.0$ & $0.6 \pm 1.2$  &  $7.5 \pm 2.6$ & $36.4 \pm 12.4$ & $0.8 \pm 1.2$ \\
& SCORE & $0.7 \pm 0.9$ & $3.6 \pm 4.7$ & $0.5 \pm 0.7$  &  $7.2 \pm 2.5$ & $41.3 \pm 7.9$ & $0.6 \pm 1.0$ \\
& CAM & $2.8 \pm 2.4$ & $19.3 \pm 16.7$ & $2.1 \pm 1.8$  &  $10.4 \pm 3.2$ & $48.8 \pm 14.8$ & $4.2 \pm 3.3$ \\
& GES & $10.5 \pm 2.5$ & $58.5 \pm 14.7$ & $-$  &  $25.6 \pm 2.3$ & $80.4 \pm 3.7$ & $-$ \\
\midrule
\multirow{4}{*}{d=20}  & NoGAM & $6.1 \pm 2.3$ & $71.5 \pm 40.0$ & $4.2 \pm 1.9$  &  $26.6 \pm 5.0$ & $237.8 \pm 21.4$ & $4.3 \pm 2.3$ \\
& SCORE &  $3.1 \pm 2.1$ & $13.2 \pm 10.2$ & $0.8 \pm 0.6$  &  $27.1 \pm 4.1$ & $204.4 \pm 23.3$ & $3.7 \pm 1.7$ \\
& CAM & $8.6 \pm 2.7$ & $100.0 \pm 38.3$ & $5.6 \pm 1.6$  &  $29.0 \pm 6.6$ & $270.7 \pm 41.7$ & $8.6 \pm 4.3$ \\
& GES & $27.1 \pm 3.1$ & $187.3 \pm 31.1$ & $-$  &  $59.8 \pm 5.1$ & $348.6 \pm 11.1$ & $-$ \\
\bottomrule
\end{tabular}
\end{table}

\begin{table}[htb!]
\footnotesize
\caption{Gauss noise SF graphs}
\label{tab:gaussian_sf}
\centering
\begin{tabular}{clcccccc}
\toprule
\multicolumn{2}{c}{} & \multicolumn{3}{l}{SF1 (sparse)} & \multicolumn{3}{l}{SF4 (dense)}\\
\midrule
 & Method & SHD & SID & $D_{top}$ & SHD & SID & $D_{top}$\\
\midrule
\multirow{4}{*}{d=10}   & NoGAM &  $0.3 \pm 0.5$ & $2.0 \pm 4.0$ & $0.2 \pm 0.4$  &  $6.4 \pm 1.9$ & $34.1 \pm 5.3$ & $0.0 \pm 0.0$ \\
& SCORE & $0.3 \pm 0.6$ & $2.7 \pm 5.8$ & $0.1 \pm 0.3$  &  $7.6 \pm 2.9$ & $31.6 \pm 9.4$ & $1.1 \pm 0.7$ \\
& CAM & $0.2 \pm 0.4$ & $1.5 \pm 3.0$ & $0.0 \pm 0.0$  &  $9.8 \pm 2.3$ & $39.7 \pm 9.9$ & $1.0 \pm 0.9$ \\
& GES & $10.6 \pm 3.0$ & $49.7 \pm 15.2$ & $-$  &  $24.1 \pm 3.3$ & $80.7 \pm 3.5$ & $-$ \\
\midrule
\multirow{4}{*}{d=20}  & NoGAM & $1.2 \pm 0.9$ & $14.2 \pm 16.3$ & $0.7 \pm 0.6$  &  $15.8 \pm 4.6$ & $224.6 \pm 22.4$ & $2.3 \pm 1.6$ \\
& SCORE & $0.9\pm 0.9$ & $13.8 \pm 12.6$ & $0.7 \pm 0.8$  &  $17.5 \pm 3.5$ & $179.2 \pm 23.8$ & $4.9 \pm 3.0$ \\
& CAM & $0.3 \pm 0.5$ & $1.9 \pm 5.7$ & $0.0 \pm 0.0$  &  $24.8 \pm 3.3$ & $240.7 \pm 29.8$ & $3.1 \pm 2.4$ \\
& GES & $28.1 \pm 7.6$ & $212.6 \pm 51.0$ & $-$  &  $58.5 \pm 3.6$ & $360.5 \pm 2.7$ & $0.0 \pm 0.0$ \\
\bottomrule
\end{tabular}
\end{table}

\begin{table}[htb!]
\footnotesize
\caption{Gumbel noise SF graphs}
\label{tab:gumbel_sf}
\centering
\begin{tabular}{clcccccc}
\toprule
\multicolumn{2}{c}{} & \multicolumn{3}{l}{SF1 (sparse)} & \multicolumn{3}{l}{SF4 (dense)}\\
\midrule
 & Method & SHD & SID & $D_{top}$ & SHD & SID & $D_{top}$\\
\midrule
\multirow{4}{*}{d=10}   & NoGAM &  $0.9 \pm 0.9$ & $6.0 \pm 5.2$ & $0.6 \pm 0.7$  &  $7.2 \pm 2.3$ & $28.1 \pm 7.5$ & $0.2 \pm 0.4$ \\
& SCORE & $1.2 \pm 1.3$ & $5.0 \pm 5.1$ & $0.7 \pm 0.6$  &  $7.6 \pm 3.2$ & $44.0 \pm 9.3$ & $0.5 \pm 0.5$ \\
& CAM & $0.4 \pm 0.5$ & $4.0 \pm 4.9$ & $0.4 \pm 0.5$  &  $8.0 \pm 3.0$ & $37.5 \pm 13.4$ & $1.5 \pm 2.3$ \\
& GES & $10.5 \pm 2.6$ & $55.1 \pm 13.0$ & $-$  &  $24.0 \pm 3.4$ & $80.3 \pm 6.1$ & $-$ \\
\midrule
\multirow{4}{*}{d=20}  & NoGAM & $2.1 \pm 1.4$ & $26.5 \pm 20.0$ & $1.5 \pm 0.9$  &  $25.3 \pm 3.7$ & $237.3 \pm 21.0$ & $2.7 \pm 2.0$ \\
& SCORE & $3.2 \pm 1.9$ & $20.0 \pm 13.6$ & $1.1 \pm 1.4$  &  $28.1 \pm 4.1$ & $212.3 \pm 19.7$ & $5.6 \pm 2.4$ \\
& CAM & $1.5 \pm 1.6$ & $13.7 \pm 16.6$ & $0.8 \pm 0.8$  &  $26.0 \pm 4.5$ & $247.0 \pm 31.6$ & $4.7 \pm 2.2$ \\
& GES & $28.0 \pm 8.7$ & $195.9 \pm 34.7$ & $-$  &  $58.9 \pm 4.8$ & $353.3 \pm 12.7$ & $-$ \\
\bottomrule
\end{tabular}
\end{table}

\begin{table}[htb!]
\footnotesize
\caption{Laplace noise SF graphs}
\label{tab:laplace_sf}
\centering
\begin{tabular}{clcccccc}
\toprule
\multicolumn{2}{c}{} & \multicolumn{3}{l}{SF1 (sparse)} & \multicolumn{3}{l}{SF4 (dense)}\\
\midrule
 & Method & SHD & SID & $D_{top}$ & SHD & SID & $D_{top}$\\
\midrule
\multirow{4}{*}{d=10}   & NoGAM & $0.1 \pm 0.3$ & $0.9 \pm 2.7$ & $0.1 \pm 0.3$  &  $8.0 \pm 2.1$ & $30.4 \pm 8.3$ & $0.5 \pm 0.9$ \\
& SCORE & $0.5 \pm 1.2$ & $1.5 \pm 3.4$ & $0.0 \pm 0.0$  &  $8.2 \pm 1.7$ & $43.6 \pm 6.5$ & $0.5 \pm 0.7$ \\
& CAM &  $0.4 \pm 0.5$ & $1.9 \pm 3.8$ & $0.1 \pm 0.3$  &  $7.0 \pm 2.2$ & $34.0 \pm 12.1$ & $1.0 \pm 1.3$ \\
& GES & $12.8 \pm 4.7$ & $60.4 \pm 19.7$ & $-$  &  $26.1 \pm 1.9$ & $82.4 \pm 2.8$ & $-$ \\
\midrule
\multirow{4}{*}{d=20}  & NoGAM & $1.7 \pm 0.9$ & $12.6 \pm 8.6$ & $0.7 \pm 0.6$  &  $25.9 \pm 3.3$ & $244.3 \pm 22.9$ & $2.5 \pm 2.3$ \\
& SCORE & $2.6 \pm 1.1$ & $16.7 \pm 15.7$ & $1.1 \pm 1.0$  &  $27.4 \pm 3.0$ & $207.6 \pm 18.4$ & $4.8 \pm 2.2$ \\
& CAM & $1.1 \pm 0.9$ & $3.9 \pm 7.8$ & $0.1 \pm 0.3$  &  $23.7 \pm 3.9$ & $227.2 \pm 15.1$ & $2.5 \pm 1.8$ \\
& GES & $29.0 \pm 6.7$ & $207.3 \pm 34.7$ & $-$  &  $58.0 \pm 6.6$ & $353.4 \pm 7.7$ & $-$ \\
\bottomrule
\end{tabular}
\end{table}

\subsection{NoGAM with linear regression}\label{app:hp_restriction}
In this section we discuss potential robustness issues of our methodology. In particular, according to Equations \eqref{eq:res_problem} and \eqref{eq:reg_score}, our algorithm requires minimization over the space of all measurable functions for the estimation of the residuals and of the score function from such residuals. It is well known that, in practice,  due to computational limitations we need to further restrict the hypothesis space of the class of functions over which we search the solution for the regression problems, and that this can induce irreducible error. Regardless, we find that the iterative identification of leaves with the $\operatorname{argmin}$ of the Mean Squared Error (as proposed in Algorithm \ref{alg:top_order}) makes NoGAM robust with respect to the error introduced by a restrictive hypothesis space. We can intuitively justify this as follow: in order to get a wrong leaf at a specific iteration, the irreducible error on the prediction of the score of each leaf needs to be larger than the total prediction error of a generic non-leaf node, which is also increased by the hypothesis space restriction. 
However, if some variables have target function much closer to the hypothesis space than others, this can induce very different irreducible errors for different variables. If such error introduced by the biased space of functions happens to be larger for leaves, it can cause mistakes in the ordering, as it could be confused with the estimation residual. Nevertheless, we argue that often this is not the case: in order to experimentally prove our claim, we run NoGAM topological ordering inference on ER4 synthetic data, replacing KernelRidge regressor with the linear model Lasso of \texttt{scikit-learn} (\cite{scikit-learn}), that we use for both residuals estimation and score prediction from the residuals. In Table \ref{tab:hp_restricted} we see that NoGAM doesn't suffer from the restriction of the search space to linear function (as in Lasso regression algorithm), despite mechanisms of the generative model being highly nonlinear: comparing the results obtained with linear and nonlinear regression, we observe that they are almost always close and comparable within error bars. 
%
\begin{table}[htb!]
\footnotesize
\caption{Experimental performance of NoGAM using KernelRidge and Lasso regression methods for the estimation of the residuals and of the score function from the residuals. (In bold we remark the regression method giving best performance.)}
\begin{minipage}{0.5\textwidth}
\label{tab:hp_restricted}
\centering
\begin{tabular}{cllc}
\toprule
 Noise & Nodes & Regression method & $D_{top}$ \\
\midrule
\multirow{4}{*}{Beta} & $d=10$ & \textbf{KernelRidge} & $0.1 \pm 0.3$ \\
& $d=10$ & Lasso & $0.2 \pm 0.4$ \\
& $d=20$ & \textbf{KernelRidge} & $0.9 \pm 0.9$ \\
& $d=20$ & Lasso & $4.6 \pm 2.9$ \\
\midrule
\multirow{4}{*}{Gamma} & $d=10$ & KernelRidge & $0.7 \pm 0.8$ \\
& $d=10$ & \textbf{Lasso} & $0.6 \pm 0.7$ \\
& $d=20$ & \textbf{KernelRidge} & $3.7 \pm 1.6$ \\
& $d=20$ & Lasso & $4.0 \pm 1.9$ \\
\midrule
\multirow{4}{*}{Gauss} & $d=10$ & KernelRidge & $0.8 \pm 0.9$ \\
& $d=10$ & \textbf{Lasso} & $0.6 \pm 0.7$ \\
& $d=20$ & \textbf{KernelRidge} & $3.1 \pm 2.0$ \\
& $d=20$ & Lasso & $3.7 \pm 1.7$ \\
\bottomrule
\end{tabular}
\end{minipage} \hfill
\begin{minipage}{0.5\textwidth}
\begin{tabular}{cllc}
\toprule
 Noise & Nodes & Method & $D_{top}$ \\
 \midrule
\multirow{4}{*}{Gumbel} & $d=10$ & \textbf{KernelRidge} & $0.6 \pm 0.7$ \\
& $d=10$ & Lasso & $0.8 \pm 1.0$ \\
& $d=20$ & KernelRidge & $4.7 \pm 2.4$ \\
& $d=20$ & \textbf{Lasso} & $3.8 \pm 2.2$ \\
\midrule
\multirow{4}{*}{Exponential} & $d=10$ & KernelRidge & $0.8 \pm 0.8$ \\
& $d=10$ & \textbf{Lasso} & $0.6 \pm 0.7$ \\
& $d=20$ & KernelRidge & $3.6 \pm 1.6$ \\
& $d=20$ & \textbf{Lasso} & $3.5 \pm 1.7$ \\
\midrule
\multirow{4}{*}{Laplace} & $d=10$ & KernelRidge & $0.8 \pm 0.7$ \\
& $d=10$ & \textbf{Lasso} & $0.7 \pm 0.6$ \\
& $d=20$ & KernelRidge & $3.9 \pm 1.5$ \\
& $d=20$ & \textbf{Lasso} & $3.1 \pm 1.5$ \\
\bottomrule
\end{tabular}
\end{minipage}
\end{table}

\end{document}